\RequirePackage{fix-cm}
\documentclass[smallcondensed]{svjour3}     % onecolumn (ditto)
\smartqed  % flush right qed marks, e.g. at end of proof
\usepackage{graphicx}
\usepackage{amsmath}
\usepackage{amssymb}
\usepackage{longtable}
\usepackage[table]{xcolor}
\usepackage{graphicx}
\usepackage{natbib}
\usepackage{array}
\usepackage{hyperref}
\newcommand{\argmax}{\operatornamewithlimits{argmax}}
\newcolumntype{P}[1]{>{\raggedright\arraybackslash}p{#1}}
\journalname{"Machine Learning"}
\begin{document}

\title{Emotion in Reinforcement Learning Agents and Robots: A Survey.
}
%\subtitle{}


\author{Thomas M. Moerland         \and
        Joost Broekens			\and
        Catholijn M. Jonker
}

\authorrunning{Moerland et al.} % if too long for running head

\institute{T.M. Moerland, J. Broekens, C.M. Jonker \at
			Delft University of Technology \\
			Mekelweg 4, 2628CD, Delft, The Netherlands \\
			\email{\{T.M.Moerland, D.J.Broekens, C.M.Jonker\}@tudelft.nl} \\						
}

\date{Received: date / Accepted: date}

\maketitle

\begin{abstract}
This article provides the first survey of computational models of emotion in reinforcement learning (RL) agents. The survey focuses on agent/robot emotions, and mostly ignores human user emotions. Emotions are recognized as functional in decision-making by influencing motivation and action selection. Therefore, computational emotion models are usually grounded in the agent's decision making architecture, of which RL is an important subclass. Studying emotions in RL-based agents is useful for three research fields. For machine learning (ML) researchers, emotion models may improve learning efficiency. For the interactive ML and human-robot interaction (HRI) community, emotions can communicate state and enhance user investment. Lastly, it allows affective modelling (AM) researchers to investigate their emotion theories in a successful AI agent class. This survey provides background on emotion theory and RL. It systematically addresses 1) from what underlying dimensions (e.g., homeostasis, appraisal) emotions can be derived and how these can be modelled in RL-agents, 2) what types of emotions have been derived from these dimensions, and 3) how these emotions may either influence the learning efficiency of the agent or be useful as social signals. We also systematically compare evaluation criteria, and draw connections to important RL sub-domains like (intrinsic) motivation and model-based RL. In short, this survey provides both a practical overview for engineers wanting to implement emotions in their RL agents, and identifies challenges and directions for future emotion-RL research. 
\end{abstract}

\section{Introduction}
This survey systematically covers the literature on computational models of emotion in reinforcement learning (RL) agents. Computational models of emotions are usually grounded in the agent decision-making architecture. In this work we focus on emotion models in a successful learning architecture: reinforcement learning, i.e. agents optimizing some reward function in a Markov Decision Process (MDP) formulation. To directly avoid confusion: the topic does not imply the agent should `learn its emotions', i.e. emotions are rather hooked on characteristics of the MDP (like value and transition functions), and may for example well persist after learning has converged. 

One may question why it is useful to study emotions in machines at all. The computational study of emotions is an example of bio-inspiration in computational science. Many important advancements in machine learning and optimization were based on biological principles, such as neural networks, evolutionary algorithms and swarm-based optimization \citep{russell1995modern}. An example encountered in this survey is homeostasis, a concept closely related to emotions, and a biological principle that led researchers to implement goal switching in RL agents. 

The study of emotions in learning agents is useful for three research fields. First, for the machine learning (ML) community, emotions may benefit learning efficiency. For example, there are important connections to the work on (intrinsically) motivated RL. Second, researchers working on interactive machine learning and human-robot interaction (HRI) may benefit from emotions to enhance both transparency (i.e. communicate agent internal state) and user empathy. Finally, from an affective modelling (AM) perspective, where emotions are mostly studied in cognitive agents, RL agents provide the general benefits of the MDP formulation: these agents require few assumptions, can be applied to a variety of tasks without much prior knowledge, and, allow for learning. This also gives AM researchers access to complex, high-dimensional test domains to evaluate emotion theories. 

Emotion is an important part of human intelligence \citep{johnson1992basic,damasio1994,baumeister2007}. On the one hand, emotion has been defined as a response to a significant stimulus - characterized by brain and body arousal and a subjective feeling - that elicits a tendency towards motivated action \citep{calvo2014,frijda1989relations}. This emphasizes the relation of emotions with motivation and action. On the other hand, emotions have also been identified as complex feedback signals used to shape behaviour \citep{baumeister2007,broekens2013}. This view emphasizes the feedback function of emotion. The common ground in both: 1) emotions are related to action selection mechanisms and 2)  emotion processing is in principle beneficial to the viability of the individual. As an illustration, \citet{damasio1994} showed that people with impaired emotional processing (due to brain damage) show failures in work and social life. These observations connecting emotions to action selection and adaptive decision-making sparked interest in the computer science community as well, mainly following the initial work by \citet{canamero1997modeling} and \citet{gadanho1998emotion}.

We wrote this survey for two reasons. First, while the topic of emotion in RL agents has received attention for nearly 20 years, it appears to fall in between the machine learning and affective modelling communities. In particular, there is no framework connecting the variety of models and implementations. Although \citet{rumbell2012emotions} compared emotion models in twelve different agents, their work does not provide a full survey of the topic, nor does it focus on agents with a learning architecture. Our main aim is to establish such a framework, hoping to bridge the communities and potentially align research agendas. As a second motivation, this survey is also useful to engineers working on social agents and robots. Emotion has an important functional role in social interaction and social robotics \citep{fong2003survey}. Our survey is also a practical guideline for engineers who wish to implement emotional functionality in their RL-based agents and robots.

As a final note, the term `reinforcement learning' may be misleading to readers from a cognitive AI or psychological background. RL may reminisce of `instrumental conditioning', with stimulus-response experiments on short time-scales. Although indeed related, RL here refers to the {\it computational} term for a successful class of algorithms solving Markov Decision Processes by sampling and learning from data. MDPs (introduced in Section \ref{rl}) provide a generic specification for short-term and long-term sequential decision-making problems with minimal assumptions. Note that many cognitive AI approaches, that usually employ a notion of `goal', are also expressible in MDP formulation by defining a sparse reward function with positive reward at the goal state. 

The structure of this review is as follows. First, Section \ref{background} provides the necessary background on emotion and reinforcement learning from psychology, neuroscience and computer science. Section \ref{methodology} discusses the survey's methodology and proposed taxonomy. Subsequently, Sections \ref{elicitation}-\ref{function} contain the main results of this survey by systematically categorizing approaches to emotion elicitation, emotion types and emotion functionality. Additionally, the comparison of evaluation criteria is presented in (Section \ref{evaluation}). The survey ends with a general discussion of our findings, highlights some important problems and indicates future directions in this field (Section \ref{discussion}).

\section{Background} \label{background}
As many papers included in this survey build upon psychological (\ref{psychology}) and neuroscientific (\ref{neuroscience}) theories of emotion, this section provides a high-level overview of these fields. Subsequently, we position our work in the computer science and machine learning community (\ref{computerscience}). We conclude these preliminaries by formally introducing computational reinforcement learning (\ref{rl}). 

\subsection{Psychology} \label{psychology}
We discuss three dominant psychological emotion theories: categorical, dimensional, and componential theories (see also \citet{lisetti2014}). 

Categorical emotion theory assumes there is a set of discrete emotions forming the `basic' emotions. These ideas are frequently inspired by the work by \citet{ekman1987universals}, who identified the cross-cultural recognition of anger, fear, joy, sadness, surprise and disgust on facial expressions. In an evolutionary perspective, each basic emotion can be considered as an elementary response pattern, or action tendency \citep{frijda1989relations}. For example, fear has the associated action tendency of avoidance, which helps the organism to survive a dangerous situation, accompanied by a negative feeling and prototypical facial expression. However, the concept of `basic' emotions remains controversial within psychology, as is reflected in the ongoing debate about which emotions should be included. The number of emotions to be included ranges from 2 to 18, see \citet{calvo2014}. 

Dimensional emotion theory \citep{russell1978evidence} assumes an underlying affective space. This space involves at least two dimensions; usually valence (i.e. positive/negative evaluation) and arousal (i.e. activation level) \citep{russell1999core}. For example, fear is a highly arousing and negative affective state. The theory was originally developed as a `Core affect' model, i.e. describing a more long-term, underlying emotional state. \citet{osgood1964measurement} orginally added dominance as a third dimension, resulting in the PAD (pleasure, arousal, dominance) model. Dimensional models have difficulty separating emotion categories such as anger and disgust, which is a common critique on the theory.

Finally, componential emotion theory, best known as cognitive appraisal theory \citep{lazarus1991cognition}, considers emotions as the results of evaluations (appraisals) of incoming stimuli according to personal relevance. Some examples of frequently occurring appraisal dimensions are valence, novelty, goal relevance, goal congruence and coping potential. Distinct emotions relate to specific patterns of appraisal activation. For example, anger is a result of evaluating a situation as harmful to one's own goals with the emotion attributed to the responsible actor and at least some feeling of power. Some well-known appraisal theories that have been a basis for computational models are the OCC model (named after the authors Ortony, Clore and Collins) \citep{ortony1990cognitive}, the component process theory of emotions (CPT) \citep{scherer2001appraisal}, and the belief-desire theory of emotions (BDTE) \citep{reisenzein2009emotional}. Although cognitive appraisal theories describe the structure of emotion well, they are limited with respect to explaining where appraisals themselves come from, what the function of emotion is in cognition and intelligence, and how they are related to evolution.

Note that the presented theories focus on different aspects of emotions. For example, appraisal theory focuses on how emotions are {\it elicited}, while categorical emotion models focus on action tendencies, i.e. the immediate {\it function} of emotions. Some consider emotions to precede action selection, while others focus on emotions as feedback signals \citep{baumeister2007}. In this survey  emotions are considered in a reward-based feedback loop, which involves both emotion elicitation and function.

\subsection{Neuroscience} \label{neuroscience}
Affective responses and their relation to behaviour and learning have also been extensively studied in neuroscience; for a survey see \citep{rolls2008orbitofrontal}. We discuss theories by LeDoux, Damasio and Rolls. The work by \citet{ledoux2003emotional} mainly focussed on the role of the amygdala in fear conditioning. LeDoux identified that incoming sensory stimuli can directly move from thalamus to amygdala, thereby bypassing the previously assumed intermediate step through the neo-cortex. As such, the work showed that emotional responses may also be elicited without neo-cortical reasoning.

\cite{damasio1994} took a different perspective on rational emotions through the `somatic marker hypothesis'. He proposes that emotions are the result of bodily sensations, which tell the organism that current sensations (i.e. events) are beneficial (e.g. pleasure) or harmful (e.g. pain). The somatic marker is therefore a signal that can be interpreted as feedback about the desirability of current and imagined situations. The somatic marker hypothesis has been interpreted in terms of RL as well \citep{dunn2006}.

Later work by Rolls shifted the attention from the amygdala to the orbito-frontal cortex (OFC) \citep{rolls2008orbitofrontal} Imaging studies have implicated the OFC in both reinforcement and affect, with direct input connections of most sensory channels (taste, olfactory, visual, touch), while projecting to several brain areas involving motor behaviour (striatum) and autonomic responses (hypothalamus) \citep{rolls2008orbitofrontal}. Also, single neuron studies have shown that visual and taste signals (the latter being a well-known primary reinforcer) converge on the same neurons \citep{rolls1994gustatory}, coined 'conditional reward neurons'. Earlier work already identified 'error neurons', which mainly respond when an expected reward is not received \citep{thorpe1983orbitofrontal}. 

Together, these theories suggest that emotions are closely linked to reward processing. These ideas are implicitly reflected in part of the reinforcement learning-based implementations in this survey. These ideas are also reflected in Rolls' evolutionary theory of emotion \citep{rolls2008orbitofrontal}, which identifies emotions as the results of primary reinforcers (like taste, affiliative touch, pain) which specify generic goals for survival and reproductive success (like food, company and body integrity). According to Rolls, emotions exclusively emerge from these goal-related events. This view is also compatible with the cognitive appraisal view that emotions are the result of stimuli being evaluated according to their goal/need relevance. However, in cognitive appraisal theory the 'goal' is defined at a different level of abstraction. 

\subsection{Computer Science} \label{computerscience}
Affective modelling is a vibrant field in computer science with active subfields \citep{calvo2014}, including work on affect detection and social signal processing \citep{vinciarelli2012,calvo2010}, computational modelling of affect in robots and virtual agents \citep{marsella2010}, and expression of emotion in robots and virtual agents \citep{ochs2015,paiva2015,lhommet2015}. Since this survey focusses on affective modelling, in particular in RL-based agents, we provide some context by discussing emotions in different agent architectures, in particular symbolic and (non-RL) machine learning-based.

One of the earliest symbolic/cognitive architectures was Velasquez' {\it Cathexis} model \citep{velasquez1998modeling}. It incorporated Ekman's six emotions in the pet robot {\it Yuppy}, which later also formed the basis for the well-known social robot {\it Kismet} \citep{breazeal2003emotion}. Several well-known symbolic architectures have also incorporated emotions, either based on categorical emotions \citep{murphy2002emotion}, somatic marker hypothesis \citep{laird2008extending}, or appraisal theories (EMIB \citep{michaud2002emib}, EMA \citep{marsella2009} and LIDA \citep{franklin2014lida}). Although symbolic/cognitive architecture approaches are capable of solving a variety of AI tasks, they are limited with respect to learning from exploration and feedback in unstructured tasks.  

In contrast, machine learning implementations focus on learning, as the agent should gradually adapt to its environment and task. The dominant research direction in this field is reinforcement learning (RL) \citep{sutton1998reinforcement}, which we formally introduce in the next section. There are however other machine learning implementations that incorporate emotions. Some examples include agents based on evolutionary neural networks \citep{parisi2010robots}, the free-energy principle \citep{joffily2013emotional}, Bayesian models \citep{antos2011using} or entropy \citep{belavkin2004relation}.

Finally, we want to stress that the focus of this review is on {\it agent} emotion, i.e. how it is elicited and may influence the agent's learning loop. A related but clearly distinct topic is how {\it human} emotion may act as a teaching signal for this loop. \citet{broekens2007emotion} showed human emotional feedback speeds up agent learning in a grid-world task compared to a baseline agent. There are a few other examples in this direction \citep{hasson2011emotions,moussa2013toward}, but in general the literature of emotion as a teaching signal is limited. Although the way in which humans actually tend to provide feedback is an active research topic \citep{thomaz2008teachable,knox2012,knox2013}, it remains a question whether emotions would be a viable channel for human feedback. We do not further pursue this discussion here, and put our focus on agent emotions in RL agents.

\subsection{Computational Reinforcement Learning} \label{rl}
Computational reinforcement learning (RL) \citep{sutton1998reinforcement,wiering2012reinforcement} is a successful approach that enables autonomous agents to learn from interaction with their environment. We adopt a Markov Decision Process (MDP) specified by the tuple: $\{\mathcal{S},\mathcal{A},T,r,\gamma \}$, where $\mathcal{S}$ denotes a set of states, $\mathcal{A} $ a set of actions, $T: \mathcal{S} \times \mathcal{A} \to P(\mathcal{S})$ denotes the transition function, $r: \mathcal{S} \times \mathcal{A} \times \mathcal{S} \to \mathbb{R}$ denotes the reward function and $\gamma \in (0,1]$ denotes a discount parameter. The goal of the agent is to find a policy $\pi: \mathcal{S}  \to P(\mathcal{A})$ that maximizes the expected (infinite-horizon) discounted return: 

\begin{align} 
Q^\pi(s,a) &= \mathbb{E}_{\pi,T} \Big \{ \sum_{t=0}^\infty \gamma^t r(s_t,a_t,s_{t+1}) |s_0=s,a_0=a \Big \} \nonumber \\ 
&= \sum_{s' \in \mathcal{S}} T(s'|s,a) \Big[ r(s,a,s') + \gamma \sum_{a' \in \mathcal{A}} \pi(s',a') Q^\pi(s',a') \Big] \label{eq1}
\end{align} 

\noindent where we explicitly write out the expectation over the (possibly) stochastic policy and transition function. The optimal value function is defined as 

\begin{equation} Q^\star(s,a) = \max_\pi Q^\pi(s,a)
\end{equation} 

\noindent from which we can derive the optimal policy 

\begin{equation} 
\pi^\star(s) = \argmax_{a \in \mathcal{A}} Q^\star(s,a)
\end{equation}

There are several approaches to learning the optimal policy. When the environmental dynamics $T(s'|s,a)$ and reward function $r(s,a,s')$ are known, we can use planning algorithms like Dynamic Programming (DP). However, in many applications the environment's dynamics are hard to determine. As an alternative, we can use sampling-based methods to {\it learn} the policy, known as reinforcement learning. 

There is a large variety of RL approaches. First, we can separate value-function methods, which try to iteratively approximate the cumulative return specified in equation (\ref{eq1}), and policy search, which tries to directly optimize some parametrized policy. Policy search shows promising results in real robotic applications \citep{kober2012reinforcement}. However, most work in RL utilizes value-function methods, on which we also focus in this survey. 

Among value-function methods we should identify model-free versus model-based approaches. In model-free RL we iteratively approximate the value-function through temporal difference (TD) learning, thereby avoiding having to learn the transition function (which is usually challenging). Well-known algorithms are Q-learning \citep{watkins1989learning}, SARSA \citep{rummery1994line} and TD($\lambda$) \citep{sutton1988learning}. The update equation for Q-learning is given by: 

\begin{equation}
Q(s,a) = Q(s,a) + \alpha \Big[ r(s,a,s') + \gamma \max_{a'} Q(s',a') - Q(s,a) \Big]
\end{equation}

\noindent where $\alpha$ specifies a learning rate. With additional criteria for the learning and exploration parameters we can show this estimation procedure converges to the optimal value function \citep{sutton1998reinforcement}.

Model-based RL \citep{hester2012learning} is a hybrid form of planning (like DP) and sampling (like TD learning). In model-based RL, we approximate the transition and reward function from the sampled experience. After acquiring knowledge of the environment, we can mix real sample experience with planning updates. We will write $M = \{\hat{T},\hat{r}\}$ to denote the estimated model. Note that a model is derived from the full agent-environment interaction history at time-point $t$, as given by $g_t = \{s_0, a_0, s_1, a_1, s_2, ... s_{t-1}, a_{t-1}, s_t\}$.

A final aspect we have not yet discussed is the nature of the reward function. Traditional RL specifications assume an external reward signal (known as an `external Critic'). However, as  argued by \citet{chentanez2004intrinsically}, in animals the reward signal is by definition derived from neuronal activations, and the Critic therefore resides inside the organism. It therefore also incorporates information from the internal environment, making all reward `internal'. \citet{singh2010intrinsically} identifies two types of internal reward: extrinsic internal and intrinsic internal (we will omit `internal' and simply use extrinsic and intrinsic from now on). Extrinsic reward is related to resources/stimuli/goals in the external world (e.g. food), possibly influenced by internal variables (e.g. sugar level). In RL terms, extrinsic reward explicitly depends on the content of the sensory information (i.e. the observed state). On the contrary, intrinsic reward is not dependent on external resources, but rather derived from the agent-environment history $g$ and current model $M$. An example of intrinsic reward in animals is curiosity. Intrinsic reward is domain-independent, i.e. curiosity is not related to any external resource, but can happen at any state (dependent on the agent history $g$). In contrast, extrinsic reward for food will never occur in domains where food does not occur. Intrinsic motivation has been identified to serve a developmental role to organisms. 

\section{Survey structure and methodology} \label{methodology}
We intended to include all research papers in which reinforcement learning and emotion play a role. We conducted a systematic Google Scholar search for 'Emotion' AND 'Reinforcement Learning' AND 'Computational', and for 'Emotion' AND 'Markov Decision Process'. We scanned all abstracts for the joint occurrence of emotion and learning in the proposed work. When in doubt, we assessed the full article to determine inclusion. Moreover, we investigated all papers citing several core papers in the field, for example, \citet{gadanho2001robot}, \citet{salichs2012new}, \citet{broekens2007affect} and \citet{marinier2008emotion}. This resulted in 52 papers included in this survey. A systematic overview of these papers can be found in Table \ref{overview1} and \ref{overview2}. 

\begin{figure}
  \centering
      \includegraphics[width = 1.0\textwidth]{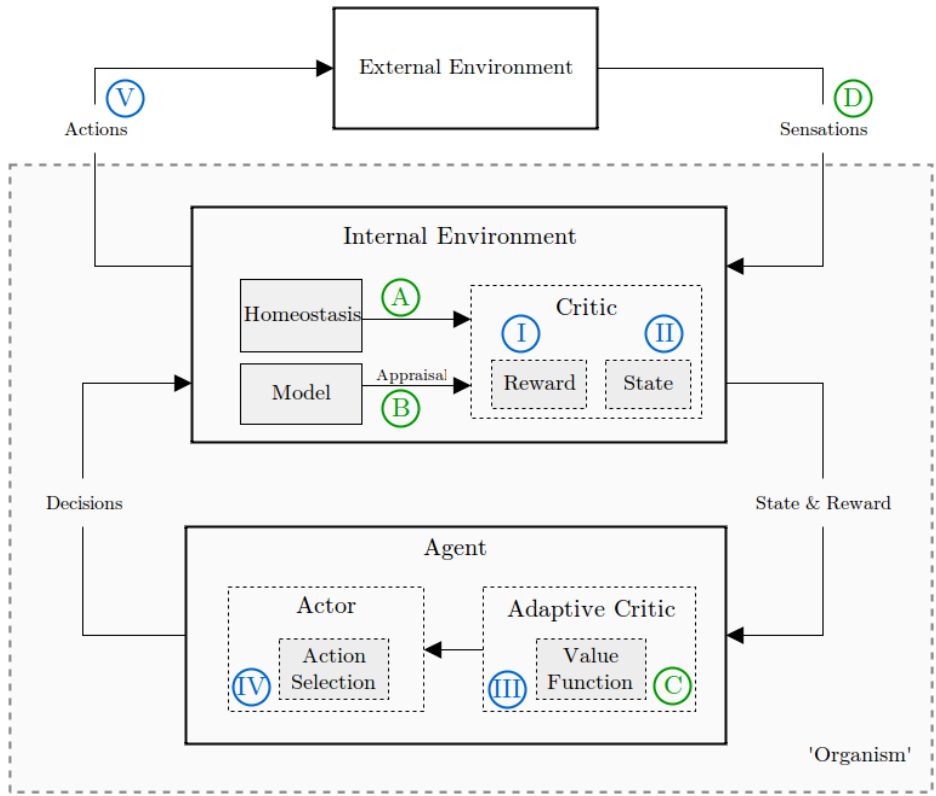}
  \caption{ Schematic representation of motivated reinforcement learning based on \citet{chentanez2004intrinsically}. Although traditional RL assumes an external Critic (to provide the reward signal), this actually happens inside the brain of real-world organisms. Thereby the Critic also incorporates, apart from external sensations, internal {\it motivations} to determine the current reward and state. Motivations have been derived from homeostatic variables and/or internal models. The Critic then feeds the state and reward to the Agent. The Agent usually learns a value function (Adaptive Critic) and determines the next action (Actor). Note that ordinary RL, in which the reward is a fully external stimulus, is still a specific case of this scheme (with the Critic as identity function). Emotion elicitation (green) has been associated to A) Homeostasis and extrinsic motivation (paragraph \ref{homeostasis}), B) Appraisal and intrinsic motivation (\ref{appraisal}), C) Reward and value function (\ref{value}) and D) Hard-wired connections from sensations (\ref{hardwired}). Subsequently, the elicited emotion may also influence the learning loop. Emotion function (blue) has been linked to I) Reward modification (\ref{reward}), II) State modification (\ref{state}), III) Meta-learning (\ref{metalearning}), IV) Action selection (\ref{actionselection}) and finally as V) Epiphenomenon (\ref{epiphenomenon}).}
  \label{schema}
\end{figure}

The proposed taxonomy of emotion elicitation, type and function is shown in Table \ref{overviewtable}, also stating the associated subsection where each category is discussed. The elicitation and function categories are also visually illustrated in Figure \ref{schema}, a figure that is based on the motivated RL illustration (with internal Critic) introduced in \citet{chentanez2004intrinsically}. Figure \ref{schema} may be useful to refer back to during reading to integrate the different ideas. Finally, for each individual paper the reader can verify the associated category of emotion elicitation, type and function through the colour coding in the overview Table \ref{overview1}.  

There is one important assumption throughout this work, which we want to emphasize here. We already introduced the distinction between extrinsic and intrinsic motivation in RL at the end of the last section. Throughout this work, we parallel extrinsic motivation with homeostasis (Section \ref{homeostasis}), and intrinsic motivation with appraisal (Section \ref{appraisal}). The extrinsic/intrinsic distinction is clearly part of the RL literature, while homeostasis and especially appraisal belong to the affective modelling literature. We group these together, as the concept of extrinsic motivation is frequently studied in combination with homeostasis, while intrinsic motivation shows large overlap with appraisal theory. We will identify this overlap in the particular sections. However, the point we want to stress is that the concepts are not synonyms. For example, it is not clear whether some intrinsic motivation or appraisal dimensions also show homeostatic dynamics (a point at which we tend to disagree with \citet{singh2010intrinsically}). However, a full discussion of the overlap and difference moves towards psychology, and is beyond the scope of our computational overview. We merely identify the overlap we observed in computational implementations, and therefore discuss both extrinsic/homeostasis and intrinsic/appraisal as single sections.  

\begin{table}[b]
\caption{\small Overview of categories in emotion elicitation, emotion type and emotion function. The number before each category identifies the paragraph where the topic is discussed. Emotion elicitation and function are also visually illustrated in Figure \ref{schema}.  } \label{overviewtable}
\small
\begin{tabular}{p{3.5cm} p{3.5cm} p{3.5cm}}
\hline\noalign{\smallskip}
\bf Emotion elicitation & \bf Emotion type & \bf Emotion function  \\
\noalign{\smallskip}\hline\noalign{\smallskip}

\ref{homeostasis} Homeostasis and \newline extrinsic motivation \newline
\ref{appraisal} Appraisal and intrinsic motivation \newline
\ref{value} Value/reward-based \newline
\ref{hardwired} Hard-wired &

\ref{categorical} Categorical \newline
\ref{dimensional} Dimensional & 

\ref{reward} Reward modification \newline 
\ref{state} State modification \newline
\ref{metalearning} Meta-learning \newline
\ref{actionselection} Action selection \newline
\ref{epiphenomenon} Epiphenomenon \\
\noalign{\smallskip}\hline
\end{tabular}
\end{table}

\section{Emotion elicitation} \label{elicitation}
We identify four major categories of emotion elicitation: extrinsic/homeostatic (\ref{homeostasis}), intrinsic/appraisal (\ref{appraisal}), value function and reward-based (\ref{value}), and finally hard-wired (\ref{hardwired}). 

\subsection{Homeostasis and extrinsic motivation} \label{homeostasis}
Several computational implementations of emotions involve homeostatic variables, drives and motivations. The notion of internal drives originates from the Drive Reduction Theory developed by \citet{hull1943principles}, which identifies drive reduction as a central cause of learning. These innate drives are also known as primary reinforcers, as their rewarding nature is hard-wired in our system (due to evolutionary benefit). An example of a homeostatic variable is energy/sugar level, which has a temporal dynamic, an associated drive when in deficit (hunger) and can be satiated by an external influence (food intake). The reader might now question why machines even need something like `hunger'. However, for a robot the current energy level shows similarity to human sugar levels (and body integrity and pain show similarity to a robot's mechanical integrity, etc.). Thereby, homeostasis is a useful concept to study in machines as well (see also the remark about bio-inspiration in the Introduction). There is a vast literature on motivated reinforcement learning, see e.g. \citet{konidaris2006adaptive} and \citet{cos2013hedonic}, mainly for its potential to naturally switch between goals. Early implementations of these ideas outside the reinforcement learning framework were by \citet{canamero1997hormonal,canamero1997modeling}.
 
\begin{figure}
  \centering
      \includegraphics[width = 0.6\textwidth]{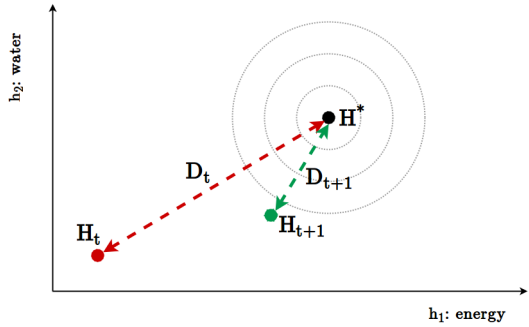}
  \caption{Schematic illustration of homeostasis and drives. The figure shows a two-dimensional homeostatic space consisting (as an example) of energy ($h_1$) and water level ($h_2$). The set point ($H^\star$) indicates the desired values for the homeostatic variables. At the current timepoint $t$ the agent's homeostatic status is $H_t$ (red). The associated drive $D_t$ can be visualize as the distance to the set point. Note that we use the Euclidean distance for the drive here (i.e. $D_t=||H^\star - H_t||_2$), while the text describes the $L_1$-norm example (i.e. $D_t=||H^\star - H_t||_1$, equation \ref{eq6}). We are free to choose any distance metric in homeostatic space. After taking an action the new homeostatic status becomes $H_{t+1}$ (green), in this case bringing both homeostatic levels closer to their set point. The difference between the drives at both timepoints has been associated to reward and joy (see Section \ref{reward}). Figure partially based on \citet{keramati2011reinforcement}.}
  \label{homeostasisfigure}
\end{figure}

We denote a homeostatic variable by $h_t$, where $t$ identifies the dependency of this variable on time. The organism's full physiological state is captured by $H_t = \{h_{1,t}, h_{2,t} .. h_{N,t}\}$, where $h_{i,t}$ indicates the $i^{th}$ homeostatic variable. Each homeostatic variable has a certain set point $H^\star = \{h^\star_1, h^\star_2 .. h^\star_N \}$ \citep{keramati2011reinforcement}. Furthermore, each homeostatic variable is affected by a set of external resources, associated to a particular action or state. For example, a particular homeostatic variable may increase upon resource consumption, and slightly decrease with every other action \citep{konidaris2006adaptive}. More formally, denoting resource consumption by $\bar{a}$ and the presence of a resource by $\bar{s}$, a simple homeostatic dynamic would be 

\begin{align}
 h_{i,t+1} =
 \begin{cases} h_{i,t} + \psi(s_t,a_t)   & \mbox{if } a_t \in  \bar{a}, s_t \in \bar{s} \\
  h_{i,t} - \epsilon    &  \mbox{otherwise} 
  \end{cases}
\end{align}

\noindent for a resource effect of size $\psi(s_t,a_t)$. We can also explicitly identify a {\it drive} as the difference between the current value and setpoint, i.e. $d_{i,t} = |h^\star_i - h_{i,t}|$ \citep{cos2013hedonic}. The overall drive of the system can then be specified by

\begin{equation}
D_t = \sum_{i=1}^N \theta_i d_{i,t} = \sum_{i=1}^N \theta_i |h^\star_i - h_{i,t}| \label{eq6}
\end{equation}

\noindent where we introduced $\theta_i$ to specify the weight or importance of the $i$-th homeostatic variable. Most examples take the absolute difference between current value and setpoint (i.e. the $L_1$ norm) as shown above. However, we can consider the space of homeostatic variables $H \in \mathbb{R}^N$ and in principle define any distance function in this space with respect to the reference point $H^\star$ (see e.g. Figure \ref{homeostasisfigure} for a Euclidean distance example). 

The weight of each homeostatic variable ($\theta_i$) does not need to be fixed in time. For example, Konidaris makes it a non-linear function of the current homeostatic level $h_{i,t}$ and a priority parameter $\rho_{i,t}$: $\theta_{i,t} = f(h_{i,t},\rho_{i,t})$. The former dependence allows priorities (i.e. rewards) to scale non-linearly with the sensory input levels (an idea reminiscent of Prospect Theory \citep{kahneman1979prospect}). The priority parameters $\rho_{i,t}$ can be estimated online, for example assigning more importance to resources which are harder to obtain (i.e. that should get priority earlier). As a final note on homeostatic RL systems, note that internal variables need to be part of the state-space as well. One can either include all homeostatic variables and learn generic Q-values, or include only the dominant drive and learn drive-specific Q-values \citep{konidaris2006adaptive}.

The connection between drives/homeostasis and emotions is partially reflected in Damasio's somatic marker hypothesis \citep{damasio1994}, stating that emotions are the result of bodily sensations. In general, we identify two ways in which homeostatic systems have been used to elicit emotions. The first elicits categorical emotions from a subset of homeostatic variables, while the second derives an overall well-being $W$ from the sum of the homeostatic dimensions. 

One of the first RL systems deriving emotions from homeostasis was by \citet{gadanho1998emotion, gadanho2001robot}. They describe an extensive set of internal variables (drives), including e.g. hunger (rises per timestep in lack of resources), pain (rises with collisions), restlessness (rises with non-progress) and temperature (rises with high motor usage). Emotions are related to these physiological variables, e.g. happiness is derived from the frequent motor use or decreasing hunger, sadness from low energy, fear from collisions (with less sensitivity if the agent is hungry or restless), and anger from high restlessness. Similar ideas are put forward by \citet{coutinho2005towards}, who specifies a more biological homeostasis: blood sugar (increases with food intake), endorphine (increases with play), energy (increases with bed rest), vascular volume (increases with water intake) and body integrity (decreases with obstacle collision). Similar examples of homeostatic emotions can be found in \citet{von2012modulating}, \citet{tanaka2004associated} and \citet{goerke2006emobot}. 

A second group of implementations first defines the overall {\it well-being} ($W$). An example of a well-being specification is

\begin{equation}
W_t = K - D_t = K - \sum_{i=1}^N \theta_i  |h^\star_{i} - h_{i,t}| \label{wellbeing}
\end{equation}

\noindent where $K$ denotes a reference value. Compared to the previous paragraph, now {\it all} internal variables (instead of subsets) are combined into a single emotion. Some papers leave the specification of well-being as their emotion \citep{gadanho2003learning}. Others actually identify the positive or negative difference in well-being as happy and unhappy \citep{salichs2012new} or `hedonic value' \citep{cos2013hedonic}. 

In conclusion, there have been numerous approaches to homeostatic systems in emotional implementations. A summary of some of the most frequently encountered homeostatic dimensions is shown in Table \ref{homeostatic_table}. Although most papers use slightly different specifications for their homeostatic dimensions, it is usually a matter of labelling that does not affect the underlying principle. Homeostatic variables provide a good way to naturally implement goal and task switching. The implementation of this functionality usually involves reward modification, which is covered in Section \ref{reward}.

\begin{table}
\caption{Overview of most frequently investigated homeostatic dimensions, their associated drive in case of deficit, and the papers in which example implementations can be found.} \label{homeostatic_table}
\small
\begin{tabular}{ l  l  p{5.7cm} }
\hline\noalign{\smallskip}
\bf Homeostasic variable & \bf Drive & \bf Papers  \\

\noalign{\smallskip}\hline\noalign{\smallskip}

Food/energy & Hunger & \citep{gadanho2001robot}, \citep{salichs2012new}, \citep{coutinho2005towards} \citep{von2012modulating} \citep{goerke2006emobot} \citep{tanaka2004associated} \\

Water level & Thirst & \citep{salichs2012new} \citep{coutinho2005towards} \\

Body integrity & Pain & \citep{gadanho2001robot} \citep{coutinho2005towards} \citep{tanaka2004associated}  \citep{lee2010mobile}  \\

Activity & Restlessness & \citep{gadanho2001robot} \citep{coutinho2005towards} \citep{von2012modulating}   \\

Energy (movement) & Sleep/tiredness & \citep{salichs2012new} \citep{coutinho2005towards} \citep{von2012modulating} \citep{goerke2006emobot} \citep{tanaka2004associated}  \\

Social interaction & Loneliness &  \citep{salichs2012new}  \\

\noalign{\smallskip}\hline

\end{tabular}
\end{table}

\subsection{Appraisal and intrinsic motivation} \label{appraisal}
Appraisal theory is an influential psychological emotion theory (see Section \ref{background}). Appraisals are domain independent elements that provide (affective) meaning to a particular stimulus. As such, they are a basis for emotion elicitation, as different combinations of appraisal dimensions have different associated emotions. Examples of appraisal dimensions are novelty, recency, control and motivational relevance. These terms of course refer to abstract cognitive concepts, but in RL literature they show a large overlap with intrinsic motivation features, being independent of a specific external resource. Instead, they are functions of the agent-environment interaction history $g$ and derived model $M$:

\begin{equation}
\zeta_{j}(s,a,s') = f_j(g,M)
\end{equation} 
 
\noindent for the $j^{th}$ appraisal variable. Note that the current state and action are actually included in $g$, but we emphasize that $f_j(\cdot)$ is not a function of the actual content of any state $s$ (see Section \ref{rl} for a discussion of the extrinsic/intrinsic distinction). Rather, $f_j(\cdot)$ computes domain-independent characteristics, like `recency' which may be derived from $g$, and `motivational relevance' which can be derived by planning over $M$.  

Intrinsic motivation is an active topic in developmental robotics \citep{oudeyer2007intrinsic}. \citet{singh2010intrinsically} shows how incorporating these dimensions as extra reward provides better task achievement compared to non-intrinsically motivated agents (see Section \ref{reward}). We discuss two implementations based on these ideas more extensively: \citet{marinier2008emotion} and \citet{sequeira2011emotion}. The work by \citet{marinier2008emotion} takes a diverse set of appraisal dimensions based on Scherer's appraisal theory \citep{scherer1999appraisal}. These include both sensory processing dimensions, like suddenness, intrinsic pleasantness and relevance, and comprehension and reasoning dimensions, like outcome probability, discrepancy from expectation, conduciveness, control and power. The implementation by \citet{sequeira2011emotion} uses a smaller subset of appraisal dimensions: novelty, relevance, valence and control. Note that these appraisal-based papers only elicit appraisal dimensions, without specifying categorical or dimensional emotions on top (see Table \ref{overview1}, i.e. appraisal papers with empty middle column).

We now highlight some appraisal implementations, both to concretize their specification in MDPs, and illustrate the differences between models. \citet{sequeira2011emotion} specifies `motivational relevance' as inversely related to the distance to the goal. If we implement a planning procedure over our model $M$ which returns an estimated distance $\hat{d}(s,s^\circ)$ to the goal node $s^\circ$ from our current node $s$, then the associated appraisal variable for motivational relevance could be \citep{sequeira2011emotion}: 

\begin{equation} 
\zeta_{relevance}(s) = \frac{1}{1 + \hat{d}(s,s^\circ)} \label{ab1}
\end{equation}

\noindent Similarly, if we denote by $c(s)$ the number of time-steps since node $s$ was last visited, then we can specify a `recency' feature as \citep{bratman2012strong}:

\begin{equation} 
\zeta_{recency}(s) = 1 - \frac{1}{c(s)} \label{ab3}
\end{equation}

\noindent This example intrinsic motivation vector $\zeta = \{\zeta_{relevance}, \zeta_{recency} \}$ is used in Section \ref{reward} to show its use in reward modification. 

There are several more specifications in intrinsic motivation RL literature that reflect appraisal dimensions. For example, \citet{hester2012intrinsically} maintain an ensemble of transition models (by stochastically adding new data to each model) and derive `model uncertainty' from the KL-divergence (as a measure of the distance between two probability distributions) between the ensemble model's predictions:

\begin{equation}
\zeta_{uncertainty}(s,a) = \sum_{i \neq j} D_{KL}\Big[ T_i(s'|s,a) \| T_j(s'|s,a) \Big]
\end{equation}

\noindent for all pairs of models $i$ and $j$ in the ensemble. As a second example from their paper, `novelty' of a state-action pair is identified from the closest $L_1$-distance to a historical observation:

\begin{equation}
\zeta_{novelty}(s,a) = \min_{<s_i,a_i> \in g}  \| \langle s,a \rangle - \langle s_i,a_i \rangle \|_1 \label{novel}
\end{equation}

\noindent Recently, \citet{houthooft2016curiosity} derive `curiosity/surprise' from the KL-divergence between the old and new transition models (i.e. after updating based on the observed transition):

\begin{equation}
\zeta_{curiosity}(s,a,s') = D_{KL} \Big[ T(\omega|g_t,a,s') \| T(\omega|g_t) \Big] \label{ab5}
\end{equation} 

\noindent where $T(\omega)$ denotes the transition model parametrized by $\omega$. Together, Equations \ref{ab1}-\ref{ab5} illustrate how intrinsic motivation and appraisal theory have modelled similar notions, and gives a short illustration of the variety of concepts that are expressible in the MDP setting.

It is also important to note that appraisal theory bears similarities to many `domain-independent' heuristics developed in the planning community \cite{russell1995modern}. These of course include heuristics without a clear psychological or biological interpretation, but we mainly emphasize the potential for cross-breeding between different research fields. For example, some appraisal theories partition novelty into three sub-elements: familiarity, suddenness and predictability \citep{gratch2014appraisal}. Each of these seem to capture different computational concepts, and such inspiration may benefit intrinsic motivation and/or planning researchers. Vice-versa, psychologist could seek for results from the RL or planning literature to develop and verify psychological theory as well.

There are several other implementations of appraisal dimensions, e.g. by \citet{yu2015emotional}, \citet{lee2010mobile}, \citet{williams2015emotion}, \citet{si2010modeling}, \citet{kim2010computational}, \citet{hasson2011emotions} and \citet{moussa2013toward}. We also encounter a few explicit social dimensions, like social fairness \citep{yu2015emotional} and social accountability \citep{si2010modeling}, although the latter for example requires some symbolic reasoning on top of the RL paradigm. This illustrates how current RL algorithms (for now) have trouble learning complex social phenomena. Some of the appraisal systems also include homeostatic variables \citep{yu2015emotional}. Both \citet{williams2015emotion} and \citet{lee2010mobile} do not mention appraisal in their paper, but their dimensions can be conceptualized as intrinsic motivation nevertheless.

In summary, some appraisal-based dimensions require cognitive reasoning, and are harder to implement. However, dimensions like novelty, motivational relevance and intrinsic pleasantness are frequently implemented (see Table \ref{appraisal1}). Table \ref{appraisal2} provides a more systematic overview of the actual connections to the RL framework. These features usually require learned transition functions, recency features or forward planning procedures over the model space, which can all be derived from the history $g$. Also note that a single concept may be interpreted in very different ways (Table \ref{appraisal2}). For example, control and power have been derived from the transitions function \citep{kim2010computational}, from the number of visits to a state \citep{sequeira2011emotion}, from a forward planning procedure \citep{si2010modeling} and from the overall success of the agent \citep{williams2015emotion}. We encounter a fundamental challenge in the field here, namely how to translate abstract cognitive concepts to explicit (broadly accepted) mathematical expressions.  

\begin{table}
\caption{ Overview of frequently investigated appraisal dimensions.} \label{appraisal1}
\begin{tabular}{p{4cm} p{7.5cm} }
\hline \noalign{\smallskip}

\bf Appraisal dimension & \bf Paper  \\

\noalign{\smallskip}\hline \noalign{\smallskip}
Novelty &  \citep{sequeira2011emotion} \citep{kim2010computational} \citep{si2010modeling} \citep{williams2015emotion} \\

Recency & \citep{marinier2008emotion}  \\

Control/Power & \citep{marinier2008emotion} \citep{sequeira2011emotion} \citep{kim2010computational} \citep{si2010modeling} \citep{williams2015emotion} \\

Motivational relevance & \citep{marinier2008emotion} \citep{sequeira2011emotion} \citep{hasson2011emotions} \citep{kim2010computational} \citep{si2010modeling} \citep{williams2015emotion} \\

Intrinsic pleasantness & \citep{marinier2008emotion} \citep{sequeira2011emotion} \citep{lee2010mobile} \\

Model uncertainty & \citep{marinier2008emotion} \citep{lee2010mobile} \citep{kim2010computational} \citep{williams2015emotion} \\

Social fairness/attachment & \citep{yu2015emotional} \citep{moussa2013toward}  \\

Social accountability &  \citep{si2010modeling} \citep{kim2010computational} \\

\noalign{\smallskip}\hline

\end{tabular}
\end{table}

\begin{table}
\caption{\small Overview of the five most frequently investigated appraisal dimensions (columns) and their specific implementations in six appraisal-based papers (rows). The cell text indicates which event causes the associated appraisal dimension to be high. Note that both \citet{williams2015emotion} and \citet{lee2010mobile} do not explicitly mention appraisal theory as their inspiration, but they do derive emotions from dimensions encountered in appraisal theory. Only the implementation of \citet{marinier2008emotion} uses direct sensory information (for control and intrinsic pleasantness), which would better fit with the hard-wired approach in Section \ref{hardwired}. All other specifications rely on (an aggregate of) the agent-environment interaction history, for example on an estimated transition model $T(s'|s,a)$.} \label{appraisal2}
\begin{tabular}{ P{1.5cm}  P{1.8cm} P{1.5cm} P{1.6cm} P{1.6cm} P{1.7cm}}
\hline \noalign{\smallskip}

 & \bf Novelty/ Suddenness & \bf Control/ power & \bf Motivational relevance & \bf Intrinsic pleasantness & \bf Model uncertainty  \\

\noalign{\smallskip}\hline \noalign{\smallskip}
\citep{kim2010computational} & Ratio of $\sum_{s'} T(s'|s,a)^2$ and $T(s'|s,a)$ & Entropy reduction by act sel. & High TD & - & Low belief $b(s)$ \& high goal distance \\

\citep{lee2010mobile} & - & - & - & Low mean travel time & Mismatch of model and obs. \\

\citep{marinier2008emotion} & High time to last state visit & Absence of obstacles & Low dist. to goal & Absence of obstacles & Low progress  \\

\citep{sequeira2011emotion} & Low \# visits to state  &  High \# visits to state &  Low dist. to goal & Current reward/ value ratio & - \\

\citep{si2010modeling} & Low $T$ of obs. transition & Low dist. to higher value state & High absolute TD & - & -  \\

\citep{williams2015emotion} & Unseen/seen ratio state-space & High success/fail ratio & Part of task finished & - & Low model accuracy \\

\noalign{\smallskip}\hline

\end{tabular}
\end{table}

\subsection{Value function and reward} \label{value}
The third branch of emotion elicitation methods in RL focusses on the value and reward functions. We can generally identify four groups: value-based, temporal difference-based, average reward-based and reward-based (Table \ref{tablevalue}).

One of the earliest approaches to sequential decision making based on emotion was by \citet{bozinovski1982self,bozinovski1996emotion}, who considered emotion to be the expected cumulative reward (i.e. the state-action value) received from taking an action in that state. Thereby, Bozinovski actually developed a precursor of Q-learning grounded in emotional ideas. Other implementations have also considered emotion as the state value. For example, \citet{matsuda2011decision} maintains a separate value function for fear, which is updated when the agent gets penalized. Recent work by \citet{jacobs2014emergent} considers the positive and negative part of the state as the hope and fear signal. Another value-based approach is by \citet{salichs2012new}, who model the fear for a particular state as the worst historical Q-value associated with that state. As such, their model remembers particular bad locations for which it should be afraid.

A second group of value function related implementations of emotions are based on the temporal difference error (TD). For Q-learning, the TD is given by

\begin{equation}
\delta = r(s,a,s') + \gamma \max_{a'} Q(s',a') - Q(s,a)
\end{equation}

There has been extensive research in neuroscience on the connection between dopamine and the TD. Following these ideas, there have also been implementations connecting happiness and unhappiness to the positive and negative TD, respectively \citep{moerland2016hope,jacobs2014emergent,lahnstein2005emotive}. Models based on the temporal difference are robust against shifting the reward function by a constant (a trait that is not shared by the models of the first group of this section). More recently, \citet{moerland2016hope} extended these ideas by deriving hope and fear signals from anticipated temporal differences (through explicit forward simulation from the current node).

Another branch of emotion derivations base themselves on the average reward. For example, \citet{broekens2007affect}, \citet{schweighofer2003meta} and \citet{hogewoning2007strategies} derive a valence from the ratio between short- and long-term average reward. \citet{shi2012artificial} also derives emotions from the temporal change in reward function, while \citet{blanchard2005imprinting} uses the average reward. Other implementations interpreted the reward ifself as the emotional signal \citep{moren2000computational,balkenius1998computational,ahn2006affective}.

In conclusion, emotions have been related to the value function, temporal difference error or direct derivative of the reward function (Table \ref{tablevalue}). Note that some implementations try to incorporate a time dimensions as well (besides only the reward or value signal), e.g. \citet{moerland2016hope}, \citet{salichs2012new} and \citet{broekens2007exploration}.

\begin{table}
\caption{ \small Overview of elicitation methods based on value and/or reward functions. Implementations are either based on the raw value function, the temporal difference error, some derivative of an average reward or from the raw reward function.} \label{tablevalue} \small
\begin{tabular}{ p{2.6cm}  p{8.7cm}}
\hline \noalign{\smallskip}

\bf Method & \bf Papers  \\
\noalign{\smallskip} \hline \noalign{\smallskip}

Value & \citep{bozinovski1982self,bozinovski1996emotion} \citep{matsuda2011decision} \citep{jacobs2014emergent} \citep{salichs2012new} \\

Temporal difference & \citep{moerland2016hope} \citep{jacobs2014emergent} \citep{lahnstein2005emotive} \\

Average reward & \citep{broekens2007affect} \citep{schweighofer2003meta} \citep{hogewoning2007strategies} \citep{shi2012artificial} \citep{blanchard2005imprinting} \\

Reward & \citep{moren2000computational,balkenius1998computational} \citep{ahn2006affective} \\

\noalign{\smallskip} \hline
\end{tabular}
\end{table}

\subsection{Hard-wired} \label{hardwired}
While all three previous groups used internal agent/robot aspects, a final category specifies hard-wired connections from sensory input to emotions. A first group of implementations use the detected emotional state of another person to influence the emotion of the agent/robot \citep{hoey2013bayesian} \citep{ficocelli2015promoting}. \citet{hasson2011emotions} uses facial expression recognition systems to detect human emotion, while \citet{kubota2010emotional} uses human speech input. Note that if these agent emotions subsequently influence agent learning, then we come very close to learning from human emotional feedback (as briefly described in Section \ref{computerscience}).

There are several other implementations that pre-specify sensation-emotion connections. In general, these approaches are less generic compared to the earlier categories. Some use for example fuzzy logic rules to connect input to emotions \citep{ayesh2004emotionally}. Another example we encountered is the previous emotional state (at $t-1$) influencing the current emotional state \citep{kubota2010emotional}. An example is the Markovian transition model between emotions in \citep{ficocelli2015promoting}, with similar ideas in \citep{zhang2009design}. This is a reasonable idea for smoother emotion dynamics, but we still categorize it as hard-wired since it does not explain how initial emotions should be generated. 
 
Finally, there is also overlap with previously described elicitation methods. For example, \citet{tsankova2002emotionally} derives an emotion (frustration) directly from the collision detector. This is very similar to some homeostatic specifications, but Tsankova does not include a body integrity or pain variable (i.e. it is therefore not a homeostatic system, but the author does make the connection between pain or non-progress and frustration). In conclusion, the hard-wired emotion elicitation does not seem to provide us any deeper understanding about emotion generation in RL agents, but the papers in this category may actually implement ideas from different elicitation methods.

%%%%%%%%%%%%%%%% CH 5 %%%%%%%%%%%%%%%%%%%%%
\section{Emotion type} \label{type}
Having discussed the methods to elicit emotions, this section discusses which types of emotions are specified. We cover both categorical (\ref{categorical}) and dimensional (\ref{dimensional}) emotion models. Note however that some appraisal theory-based papers only elicit appraisal dimensions, without specifically identifying emotions (see Table \ref{overview1}). 

\subsection{Categorical} \label{categorical}
Most papers in the emotion and RL literature elicit categorical emotions. An overview of the most occurring emotions and their associated papers is presented in Table \ref{cattable}. Joy (or happiness) is the most implemented emotion by a wide variety of authors. We did not include the papers that specify a valence dimension (see Section \ref{dimensional}), but this could also be interpreted as a happy-sad dimension. A few papers \citep{von2012modulating} \citep{tanaka2004associated} specifically address Ekman's six universal emotions (happy, sad, fear, anger, surprise, disgust), while most papers drop the latter two emotions. 

In general, happy, sad, fear and anger have been implemented in all elicitation categories (homeostatic, appraisal and value-based). However, hope has mainly been connected to value function based systems. The implementations of hope try to assess anticipation (by addressing the value function \citep{jacobs2014emergent}, the dynamics within a decision cycle \citep{lahnstein2005emotive}, or explicitly forward simulating from the current node towards expected temporal differences \citep{moerland2016hope}). Hope therefore needs a time component, a notion which is not directly available from for example an extrinsic homeostasis dimension.

An overview of the most often elicited emotions (happy, sad, fear and angry) is provided in Table \ref{cattable2}. The table shows that different elicitation methods have been associated to similar sets of categorical emotions. For example, anger (fourth column) has been associated to extrinsic homeostasis (e.g. hunger), intrinsic appraisal (e.g. non-progress) and reward-based (decreasing received reward) elicitation. Note that frustration, a closely related emotion, has been associated to obstacle detection \citep{tsankova2002emotionally} and non-progress \citep{hasson2011emotions} as well. The other three emotions in Table \ref{cattable2} have also been associated to each elicitation dimension, as is easily observed from the colour coding. 

Note that Table \ref{cattable2} also shows how different researchers apply different elicitation methods within one paper (i.e. looking at rows instead of columns now). Moreover, a few papers even combine elicitation methods for an individual emotion. For example, \citet{williams2015emotion} derives fear from a combination of pain (extrinsic) and novelty (intrinsic/appraisal). It is important to realize that the elicitation methods of the previous section are clearly only a framework. These are not hard separations, and combining different approaches is clearly possible (and probably necessary), as these papers nicely illustrate. 

Finally, many included papers did not fully specify the implemented connections between elicitation method and emotion type, making it difficult to replicate these studies. For example, \citet{von2012modulating} only mentions the connections between homeostatic dimensions and emotions are based on fuzzy logic, but does not indicate any principles underlying the real implementation. Similar problems occur in \citep{tanaka2004associated}, \citep{ayesh2004emotionally} and \citep{obayashi2012emotional}, while \citep{zhou2002computational} and \citep{shibata1997artificial} leave the implemented connections unspecified.   

\begin{table}
\caption{\small Overview of categorical emotion implementations.} \label{cattable}
\begin{tabular}{ p{2.7cm}  p{8.6cm} } 
\hline \noalign{\smallskip}
\bf Categorical emotion & \bf Paper  \\

\noalign{\smallskip} \hline \noalign{\smallskip}
Joy/happy &  \citep{gadanho2001robot} \citep{von2012modulating} \citep{ficocelli2015promoting} \citep{tanaka2004associated} \citep{goerke2006emobot} \citep{yu2015emotional} \citep{lee2010mobile} \citep{williams2015emotion} \citep{hasson2011emotions} \citep{moussa2013toward} \citep{salichs2012new} \citep{cos2013hedonic} \citep{moerland2016hope} \citep{jacobs2014emergent} \citep{lahnstein2005emotive} \citep{shi2012artificial} \citep{el2000flame} \citep{kubota2010emotional} \\

Sad/unhappy/distress &  \citep{gadanho2001robot} \citep{von2012modulating} \citep{ficocelli2015promoting} \citep{tanaka2004associated} \citep{yu2015emotional} \citep{lee2010mobile} \citep{moussa2013toward} \citep{salichs2012new} \citep{moerland2016hope} \citep{jacobs2014emergent} \citep{lahnstein2005emotive} \citep{el2000flame} \citep{kubota2010emotional} \\

Fear &  \citep{gadanho2001robot} \citep{von2012modulating} \citep{tanaka2004associated} \citep{goerke2006emobot} \citep{yu2015emotional} \citep{lee2010mobile} \citep{williams2015emotion} \citep{salichs2012new} \citep{moerland2016hope} \citep{jacobs2014emergent} \citep{matsuda2011decision} \citep{shi2012artificial} \citep{el2000flame} \citep{kubota2010emotional} \\

Anger & \citep{gadanho2001robot} \citep{von2012modulating} \citep{ficocelli2015promoting} \citep{tanaka2004associated} \citep{goerke2006emobot} \citep{yu2015emotional} \citep{hasson2011emotions} \citep{moussa2013toward} \citep{shi2012artificial} \citep{el2000flame} \citep{kubota2010emotional} \\

Surprise & \citep{von2012modulating} \citep{tanaka2004associated} \citep{lee2010mobile} \\

Hope & \citep{moerland2016hope} \citep{jacobs2014emergent} \citep{lahnstein2005emotive} \citep{el2000flame}  \\

Frustration & \citep{hasson2011emotions} \citep{huang2012goal} \citep{tsankova2002emotionally} \\

\noalign{\smallskip} \hline
\end{tabular}
\end{table}

\begin{table}
\caption{\small Overview of four categorical emotion (columns) elicitations for different papers (rows). The text in each cell specifies the elicitation condition. We observe different categories of emotion elicitation, i.e. homeostatic (blue, Section \ref{homeostasis}), appraisal (green, \ref{appraisal}) and value-based (red, \ref{value}). We see how single emotions are connected to different elicitation methods (multiple colours in single column) and how single papers use different elicitation methods (multiple colours in single row).} \label{cattable2}
\begin{tabular}{ P{2cm}  P{2cm} P{2cm}  P{2cm}  P{2cm}  }
\hline \noalign{\smallskip}

 & \bf Happy/Joy  & \bf Sad/Distress & \bf Fear & \bf Anger  \\

\noalign{\smallskip} \hline
 \citep{gadanho1998emotion} & High energy \cellcolor{blue!20} & Low energy \cellcolor{blue!20} & Pain \cellcolor{blue!20} & High restlessness (low progress) \cellcolor{green!20} \\ 
 
 \citep{goerke2006emobot}& All drives low \cellcolor{blue!20} & - \cellcolor{black!8}  & Homesick \& low energy \cellcolor{blue!20} & Hunger \& homesick \& high energy \cellcolor{blue!20} \\
 
 \citep{kim2010computational} & Goal achievement \newline \cellcolor{green!20} & No goal achievement \cellcolor{green!20} & Pain \cellcolor{blue!20} & No progress \cellcolor{green!20} \\ 
 
 \citep{williams2015emotion} & Progress \& control \& low pain \cellcolor{green!20} & - \cellcolor{black!8}  & Pain \& novelty \cellcolor{green!20} & - \cellcolor{black!8} \\
 
 \citep{salichs2012new} & Positive delta well-being \newline \cellcolor{blue!20} & Negative delta well-being \cellcolor{blue!20} & Worst historical Q(s,a) \cellcolor{red!20} & - \cellcolor{black!8} \\ 
 
 \citep{moerland2016hope} & Positive TD \cellcolor{red!20} & Negative TD \cellcolor{red!20} & Anticipated negative TD \newline \cellcolor{red!20} &  - \cellcolor{black!8} \\
 
 \citep{shi2012artificial} & Increasing positive reward \newline \cellcolor{red!20} & - \cellcolor{black!8}  & Increasing negative reward \cellcolor{red!20} & Decreasing positive reward \cellcolor{red!20} \\
 
 \citep{yu2015emotional} & High well-being \cellcolor{blue!20} & Egoistic agent \& low well-being \cellcolor{blue!20}  & Agent defects \& others co-orperate \cellcolor{green!20} & Agent co-orperates \& others defect \cellcolor{green!20} \\

\hline

\end{tabular}
\end{table}

\subsection{Dimensional} \label{dimensional}
Relative to the number of implementations of categorical emotions, there is a much smaller corpus of work on dimensional emotions (Table \ref{dimtable}). The most implemented dimension is valence.  Not surprisingly, valence has mostly been derived from reward-based elicitation methods \citep{broekens2007affect} \citep{ahn2006affective} \citep{zhang2009design} \citep{obayashi2012emotional} \citep{hogewoning2007strategies}. It is also connected to a few extrinsic homeostasis papers \citep{coutinho2005towards} \citep{gadanho2003learning}, but then it is referred to as `well-being'. Although this is not completely the same concept, we group these together here for clarity. 

Following the dimensional emotion models of \citet{russell1999core} introduced in Section \ref{psychology}, the second most implemented dimension is arousal. Arousal has been connected to extrinsic homeostatic dimensions (e.g. pain and overall well-being \citep{coutinho2005towards}), appraisal-like dimensions (e.g. continuation of incoming stimulus \citep{kuremoto2013improved}), and a few hard-wired implementations \citep{ayesh2004emotionally} \citep{guojiang2010behavior}. Note that some do not use the term arousal but refer to similar concepts, e.g. relaxation \citep{coutinho2005towards} and restlessness \citep{ayesh2004emotionally}. The only paper to extend the valence-arousal space is by \citet{hoey2013bayesian}, who also include control. 

In general, the dimensional emotion models seem somewhat under-represented compared to the categorical emotion implementations. Although the implementation for valence shows some consistency among papers, there is more difficulty to specify arousal or different emotion dimensions. Nevertheless, the continuous nature of dimensional emotion models remains appealing from an engineering perspective. A possible benefit is the identification of a desirable target area in affective space, towards which the agent aims to progress \citep{guojiang2010behavior}. 

\begin{table}
\caption{\small Overview of dimensional emotion implementations.} \label{dimtable} \small
\begin{tabular}{p{1.7cm} p{9.5cm}}
\hline\noalign{\smallskip}
\bf Dimensional emotion & \bf Paper  \\
\noalign{\smallskip}\hline\noalign{\smallskip}

Valence & \citep{kuremoto2013improved} \citep{ahn2006affective} \citep{zhang2009design} \citep{broekens2007affect} \citep{broekens2007emotion} \citep{obayashi2012emotional} \citep{hogewoning2007strategies} \citep{hoey2013bayesian} \citep{guojiang2010behavior} \citep{coutinho2005towards}  \\

Arousal & \citep{kuremoto2013improved} \citep{obayashi2012emotional} \citep{ayesh2004emotionally} \citep{hoey2013bayesian} \citep{guojiang2010behavior} \citep{coutinho2005towards} \\

Control & \citep{hoey2013bayesian}\\

\hline\noalign{\smallskip}
\end{tabular}
\end{table}

%%%%%%%%%% CH 6 %%%%%%%%%%%%%%%%
\section{Emotion function} \label{function}
We now discuss the ways in which emotions may influence the learning loop. It turns out emotions have been implicated with all main aspects of this loop: Reward (\ref{reward}), State (\ref{state}), Adaptive Critic (\ref{metalearning}) and Actor (\ref{actionselection}). Finally, emotion has also been studied as an epiphenomenon, i.e. without any effect on the learning loop, but for example to communicate the learning/behavioural process to other social companions (\ref{epiphenomenon}). These categories are visualized in Figure \ref{schema} (labels I-V). Note that this Section introduces the ways in which emotion may influence the RL loop on a conceptual level. We summarize the resulting effect, for example on learning efficiency, in Section \ref{evaluation}. 

%%%
\subsection{Reward modification} \label{reward}
A large group of emotional RL implementations use emotions to modify the reward function. These approaches add an additive term to the reward function that relies on emotions (we have only encountered additive specifications). The reward function is given by

\begin{equation}
r_t = \tilde{r}_t + r^\triangle_t \label{rewsha}
\end{equation}

\noindent where $\tilde{r}(t)$ denotes the external reward function and $r^\triangle(t)$ an internal reward based on emotional mechanisms. In the RL community, Eq. \ref{rewsha} is known as {\it reward shaping} \citep{ng1999policy}. The internal reward can be targeted at maximizing positive emotions, but is also frequently associated to homeostatic variables or appraisal dimensions (see Sections \ref{homeostasis} and \ref{appraisal} for elicitation). However, the general underlying principle usually remains that agents seek to maximize positive emotions and minimize negative emotions.

\paragraph{Homeostasis.}
For homeostatic systems the reward becomes dependent on the current state of the internal homeostatic variables. Some implementations use the difference in overall well-being,

\begin{equation}
r^{\triangle}_t = W_t - W_{t-1} = D_{t-1} - D_{t}
\end{equation}

\noindent where the step from well-being $W$ to overall drive $D$ naturally follows from Equation (\ref{wellbeing}). In this specification, the acquisition of food does not provide any reward if the associated homeostatic variable (e.g. energy/sugar level) is already satiated. Implementations of the above idea can be found in \citep{gadanho2001robot}, \citep{salichs2012new} and \citep{cos2013hedonic}. Variants of this have focussed on using positive emotions (instead of well-being) as the reinforcement learning signal, e.g. in \citep{gadanho1998emotion} and \citep{goerke2006emobot}.

\paragraph{Appraisal-based.}
Similar ideas are used for appraisal-based reward modifications. Some examples of appraisal dimension specifications were discussed in Section \ref{appraisal}, with some formal examples in Equations \ref{ab1}-\ref{ab5}. Appraisal dimensions are related to generic concepts of the agent history (novelty, recency, consistency of observations with world model) and expectations with respect to the goal (motivational relevance, intrinsic pleasantness). Several studies in the intrinsically motivated reinforcement learning literature have identified the learning and survival benefit of these dimensions \citep{oudeyer2007intrinsic,oudeyer2007intrinsic2}. Some authors therefore took appraisal theory as an inspiration to develop intrinsic motivation features. 

Specifications in this direction therefore usually take the following form:

\begin{equation}
r^{\triangle}_t = \sum_{j=1}^J \phi_j \zeta_{j}(g_t)
\end{equation}

\noindent for $J$ appraisal variables and $\phi_j$ denoting the weight of the $j$-th appraisal dimension. We could for example use the two features in Equations \ref{ab1}-\ref{ab3}, specifying an agent that gets rewarded for motivational relevance and recency. Note that appraisal specifications usually do not include the difference with $(t-1)$, probably because they are usually assumed not to satiate (i.e. no underlying homeostatic dynamics). We also note that a reward bonus for novelty (e.g. as in Eq. \ref{novel}) is in the RL literature usually referred to as `optimism in the face of uncertainty', i.e. we want to explore where we have not been yet. 
 
\citet{sequeira2011emotion} actually tries to optimize the vector of weights $\phi$ (with respect to overall goal achievement). In a more recent publication, \citet{sequeira2014learning} also extends this work to actually learn the required appraisal dimensions through genetic programming. Similar ideas can be found in \citet{marinier2008emotion}. One of the problems with both implementations is the distance-to-goal heuristic used by both emotion-based agents, which has access to additional information compared to the baseline agent (although the heuristic does not monotonically increase with the actual distance to goal). We discuss the empirical results of these papers more systematically in Section \ref{evaluation}. 

%%%
\subsection{State modification} \label{state}
Emotions have also been used as part of the state-space (learning emotion specific value functions and policies). An example is the social robot Maggie \citep{castro2013autonomous}. When fear is elicited it becomes part of the state-space (replacing the dominant drive in a homeostatic system), which makes Maggie learn fear-specific action values. 

Some papers explicitly write $Q(s,a,e)$, where $e$ denotes the emotional state, to illustrate this dependency \citep{ahn2006affective} \citep{ayesh2004emotionally}. More examples of such implementations can be found in \citep{zhang2009design} \citep{ficocelli2015promoting} \citep{obayashi2012emotional} and \citep{matsuda2011decision}. Hoey developed a POMDP variant called Bayesian Affect Control Theory that includes the three-dimensional emotional space (valence, control, arousal) of a companion \citep{hoey2013bayesian} and the agent itself \citep{hoey2015bayesian}. There are also implementations that use reinforcement learning to model the affective state of a human or group \citep{kimgroup2015}, but note that this is a different setting (i.e. RL to steer human emotional state instead of agent emotional state).

Using emotion to modify the state can also be seen as a form of representation learning. There are not many architectures that learn the modification (most hard-code the emotion elicitation), with the exception of \citet{williams2015emotion}. Their architecture has similarities to the bottle-neck structure frequently encountered in deep neural network research, for example in (deep) auto-encoders \citep{goodfellow2016deep}. We return to the fully-learned approach in the Discussion (Section \ref{discussion}). 

\subsection{Meta-learning} \label{metalearning}
The previous two sections showed how emotion has been implicated with determining both the reward and state, which together can be considered as the (Internal) Critic. Afterwards, the state and reward are used to learn a value function, a process that is usually referred to as the Adaptive Critic (see Figure \ref{schema}). The learning process requires appropriate (and tedious) scaling of learning parameters, most noteworthy the learning rate $\alpha$ (see Section \ref{rl}). 

The connection between emotion and these learning parameters was inspired by the work of \citet{doya2000metalearning,doya2002metalearning}. He identified neuroscientific grounding for the connection between several neurotransmitters and several reinforcement learning parameters. In particular, he proposed connections between dopamine and the temporal difference error ($\delta$), serotonin and the discount factor ($\gamma$), noradrenaline and the Boltzmann action selection temperature ($\beta$) and acetylcholine and the learning rate ($\alpha$). 

This work inspired both \citet{shi2012artificial} and \citet{von2012modulating} to implement emotional systems influencing these metaparameters. Shi identifies the connections joy $\to \delta$, anger $ \to \beta$, fear $\to \alpha$ and relief $\to \gamma$, while von Haugwitz changes only the latter two to surpise $ \to (1-\alpha)$ and fear $ \to (1-\gamma)$. 

Recently, \citet{williams2015emotion} also investigated metaparameter steering in navigation tasks. Together with \citep{sequeira2014learning} they are the only ones to {\it learn} the emotional connections, and then post-characterize the emerged phenomena. Williams trains a classifier connecting a set of primary reinforcers (both appraisal and homeostasis-based) to the metaparameters of their navigation algorithm. They train two emotional nodes, and only afterwards anthropomorphicized these. One node learned positive connections to progress and control and negatively to pain and uncertainty, while it caused the robot to increase its speed and reduce the local cost bias. In contrary, their second node was elicited by pain and novelty, while it caused the opposite effect of node 1. They afterwards characterized these nodes as `happy' and `fear', respectively. 

\subsection{Action selection} \label{actionselection}
The final step of the RL loop involves action selection. This incorporates another crucial RL challenge, being the exploration/exploitation trade-off. Emotions have long been implicated with action readiness, and we actually already encountered two papers steering the Boltzmann action selection temperature $\beta$ above (as it is technically also a metaparameter of the RL system). We next focus on those papers that specifically target action selection. 

One branch of research focusses on directly modifying the exploration parameter. \citet{broekens2007exploration,broekens2007affect} has done extensive investigations of the connections between valence and the exploration/exploitation trade-off. In one implementation \citep{broekens2007affect} selection was based on internal simulation, where a valency determined the threshold for the simulation depth. In another paper \citep{broekens2007exploration} this valency directly influenced the $\beta$ parameter in a Boltzmann action selection mechanism. \citet{schweighofer2003meta} applied small perturbations to the exploration parameters based on emotion, and subsequently kept the parameters if they performed better. Finally, \citet{hogewoning2007strategies} investigated a hybrid system of Broekens and Schweighofer, trying to combine their strengths. 

Other papers use emotion to switch between multiple sets of value functions, thereby effectively determining which set should currently be used for action selection. For example, both \citet{tsankova2002emotionally} and \citet{hasson2011emotions} use a high frustration to switch between behaviour. Similarly, \citet{kubota2010emotional} use several emotions to switch between the weighting of different value functions. For example, happiness leads to exploration by selecting a value function derived from inverse recency. Note that such a recency feature was used in the appraisal section described previously, but there it modified the reward function, while now emotion is used to switch between value functions. Although this technically leads to similar behaviour, emotion intervenes at a different level. 

\subsection{Epiphenomenon} \label{epiphenomenon}
The final category of functions of emotions seems an empty one: Epiphenomenon. Several papers have studied emotion elicitation in RL, without the emotion influencing the learning or behavioural loop. These papers usually focus on different evaluation criteria as well (see Section \ref{evaluation}). Examples of papers that only elicit emotions are \citep{coutinho2005towards}, \citep{goerke2006emobot}, \citep{si2010modeling}, \citep{kim2010computational}, \citep{bozinovski1982self,bozinovski1996emotion}, \citep{jacobs2014emergent}, \citep{lahnstein2005emotive} and \citep{moerland2016hope}.

There can however still be a clear function of the emotion for the agent in a social communication perspective (node V in Figure \ref{schema}). Emotion may communicate the current learning and behavioural process, and also create empathy and user investment. The potential of emotions to communicate internal state and enhance empathy is infrequently evaluated in current reinforcement learning related emotion literature. This seems a fruitful direction when emotions serve to make an agent or robot more sociable and likeable. 

This concludes our discussion of emotion functions in RL agents. The full overview is provided in Table \ref{overview2}, which mainly lists the categories per paper. The most important connections between Sections \ref{elicitation}-\ref{function} (i.e. column 1 to 3 in Table \ref{overview1}) were described in the text and tables (e.g. Table \ref{appraisal2} and \ref{cattable2}). 

%%%%% CH 7

\section{Evaluation} \label{evaluation}
This section systematically addresses the embodiment, test scenario and main empirical results found in the different papers. A systematic overview of this section is provided in Table \ref{overview2}.

\subsection{Embodiment}
We can grossly identify 5 embodiment categories: standard single agent, multiple agents, screen agents, simulated robot and real robot. The standard agent setting usually concerns a (gridworld) navigation simulation in some environment designed by the researcher. Some agents are also designed to appear on a screen for interaction with a user \citep{el2000flame}. Another group of embodiments concern simulated or real robots. Simulated robots are based on models of existing real robots, i.e. they usually incorporate more realistic physics and continuous controls. 

There are also real robotic implementations in navigation and resource tasks. However, several robotic implementations (especially those involving human interaction) use the robot mainly as physical embodiment (without moving much, for example in a dialogue task). Overall, most implementations have focussed on simulated agents. It is important to note that most state-spaces stay relatively small, i.e. sensory information usually has narrow bandwidth (or is assumed to be appropriately pre-processed). Although this facilitates interpretation, a remaining question is whether the current emotion modelling methods scale to high-dimensional and complex problems. 

\subsection{Test scenario}
Emotion implementations have been tested in different scenarios: navigation tasks with resources and/or obstacles, multiple agent interaction settings and human-agent/robot interaction tasks. 

There is a wide variety of navigation tasks with additional (multiple) resources and obstacles (with associated positive and negative rewards). When resources and obstacles are non-stationary we usually see the terminology `prey' and `predators'. Within this group we mainly see navigation tasks with a single goal and multiple obstacles (i.e. `mazes' \citep{marinier2008emotion} or robot navigation \citep{lee2010mobile} \citep{williams2015emotion}). A second group involves multiple resources, which are mostly connected to underlying homeostatic systems to investigate behaviour switching. A few tasks also specifically include virtual enemies \citep{sequeira2011emotion} or humans with adversarial intentions \citep{castro2013autonomous} \citep{tanaka2004associated}. 

A second, much smaller group of scenarios involves multiple agents in a social simulation scenario, either a competitive \citep{von2012modulating} \citep{yu2015emotional} or co-operative one \citep{matsuda2011decision}. The third category tests their implementation in interaction with humans. This can either involve a human dialogue task \citep{ficocelli2015promoting} \citep{moussa2013toward} or physical interaction with a human \citep{blanchard2005imprinting} \citep{shibata1997artificial}. 

In general, most papers have constructed their own scenario. We have not seen any test scenarios being borrowed from other emotion-learning implementations, nor from the general reinforcement learning literature. This makes it hard to compare different implementations amongst each other.  

\subsection{Main results}
Finally, we discuss what empirical results were found by the various authors. We identify three main categories in which emotions may be useful to the agent: learning efficiency, emotion dynamics and human-robot interaction (HRI) (Table \ref{overview2}, third column). 

\paragraph{Learning efficiency.} Most authors in emotion-RL research have focussed on learning efficiency (see Table \ref{overview2}). Overall, emotions have been found beneficial in a variety of learning tasks. Agents with emotional functionality achieved higher average rewards \citep{gadanho2001robot,sequeira2014learning,yu2015emotional} or learned faster \citep{marinier2008emotion,ahn2006affective,zhang2009design}. Others researchers focussed on the ability to avoid specific negative rewards, like the ability to avoid collisions \citep{gadanho2001robot,lee2010mobile} and navigate away from obstacles \citep{shi2012artificial}. Other researchers report improved behaviour switching, where emotional agents better alternate between goals \citep{cos2013hedonic,hasson2011emotions,goerke2006emobot}. Finally, some authors specifically show improved exploration \citep{broekens2007exploration}. Many authors that focussed on learning performance do compare to a non-emotional baseline agent, which is of course a necessary comparison. Altogether, the results show emotions may be a useful inspiration to improve learning performance of RL agents. 

\paragraph{Emotion dynamics.} A second group of researchers focusses on emotion dynamics, usually comparing the emergent emotion signals to known psychological theories. For example, \citet{jacobs2014emergent} showed patterns of habituation and extinction, \citet{moren2000computational} reproduced blocking, while \citet{blanchard2005imprinting} observed approach and avoidance behaviour in their emotional agent. Other researchers qualitatively interpret whether the emotion dynamics fit the (social) interaction \citep{tanaka2004associated} \citep{moussa2013toward} or occurs at appropriate states in the scenario \citep{moerland2016hope}. Altogether, results in this category show that emotion in RL agents might be a viable tool to study emotion theories in computational settings.  

\paragraph{Human-robot interaction.} Finally, a third group of researchers focusses on human-robot interaction evaluation. Their primary focus is to show how emotions may benefit social interaction with humans, usually by taking questionnaires with the participants after the experiment. Participants of \citet{ficocelli2015promoting} report more effective communication, participants of \citet{el2000flame} found the agent more convincing, and participants of \citet{shibata1997artificial} report an increased notion of connection as well as increased perception of robot intelligence. \citet{kim2010computational} describe an enhanced pleasant feeling of the participant after the human-agent interaction. Therefore, there is clear indication that emotion in RL agents may benefit an interactive learning setting. However, there are relatively few papers in this category compared to the other two, and this may be a direction for more research.

\section{Discussion} \label{discussion}
This article surveyed the available work on emotion and reinforcement learning in agents and robots, by systematically categorizing emotion elicitation, type and function in RL agents. We first summarize the main results and identify the challenges encountered throughout the article.

Emotions have been elicited from extrinsic motivation (in combination with homeostasis), intrinsic motivation (in combination with appraisal), value and reward functions and as hard-wired implementation. We want to emphasize again that extrinsic motivation and homeostasis are not synonyms, nor are intrinsic motivations and appraisal (see Section \ref{methodology}). The hard-wired emotion elicitation seems least useful, as it does not provide any deeper understanding about emotion generation, and is by definition hand-crafted to the task. The other three elicitation methods are useful and appear to address different aspects of emotions. Homeostasis focusses on the inner resource status, appraisal on the inner model status and value/reward focusses on the learning process. They seem to cover different aspects of emotions. For example, surprise seems only elicitable from a model, joy from food requires extrinsic motivation and homeostasis, while aspects like anticipated change need value functions. Finally, note that there remains slight overlap among categories, i.e. they serve as a framework, but are not mutually exclusive. This is also illustrated by the overlap among implementations in Table \ref{cattable2}.  

Regarding emotion types we observed a relatively larger corpus of categorical implementations than dimensional models. Although dimensional models are appealing from an engineering perspective, they are usually implemented in 1D (valence) or 2D (valence-arousal) space. This makes it challenging to implement a diverse set of emotions. We do want to present a hypothesis here: dimensional and categorical emotions may fit into the same framework, but at different levels. Concepts like `well-being', as encountered throughout this survey, do not appear to be categorical emotions, but could be interpreted as valence. However, an agent can have categorical emotions on top of a well-being/valence system, joining both emotion types in one system. Similarly, arousal could be related to the speed of processing of the RL loop, also entering the RL process at a different level.  

Finally, emotion function could involve nearly every node in the RL loop: reward, state, value function and action selection. It seems like all approaches are useful, as each element targets a different RL challenge. The fifth emotion function category (epiphenomenon) should get more attention because it involves a different kind of usefulness (communicative). Although quite some papers are focussing on emotion dynamics, there is less work on evaluating the potential of emotions to communicate the learning process. \citet{thomaz2006teachable} found that transparency of the learner's internal process (in their case through the robot's gaze direction) can improve the human's teaching. We hypothesize emotional communication to express internal state may serve a similar role, which is a topic that could get more research attention in the future.

\paragraph{Advice for implementation.}
We expect this article is useful to engineers who want to implement emotional functionality in their RL-based agent or robot. We advise to first consider what type of functionality is desired. When the goal is to have emotions visualize agent state, or have believable emotions to enhance empathy and user investment, then emotions can be implemented as an epiphenomenon (i.e. focus on Sections \ref{elicitation} and \ref{type}). The reader could for example first decide on the desired emotion types, and then check which available elicitation methods seem applicable (e.g. via Table \ref{overview1}). When one desires emotion function in their agent/robot as well, then Section \ref{function} becomes relevant. We advise the reader to first consider the desired functionality, e.g. a more adaptive reward function, learning parameter tuning, or modulated exploration, and then work `backwards' to emotion type and emotion elicitation. Readers may verify whether there are existing implementations of their requirements through the colour coding in Table \ref{overview1}.

In general, we believe researchers in the field should start focussing on integrating approaches. This survey intended to provide a framework and categorization of emotion elicitation and function, but it seems likely that these categories actually jointly occur in the behavioural loop. We look forward to systems that integrate multiple approaches. Moreover, we want to emphasize the paper by \citet{williams2015emotion} that took a fully learned approach. Their system contains nodes that were trained for their functional benefit, and later on characterized for the emotion patterns. We expect such an approach to both be more robust against the complexity problems encountered when developing integrated systems, and to transfer more easily between problem settings as well. 

\paragraph{Testing and quality of the field.}
We also systematically categorized the testing scenarios and evaluation criteria (Section \ref{evaluation} and Table \ref{overview2}). There are several points to be noted about the current testing. First we want to stress a point already made by \citet{canamero2003designing}, who noted that `one should not put more emotion in the agent than what is required by the complexity of the system-environment interaction'. Many of the current implementations design their own (grid) world. While these artificial worlds are usually well-suited to assess optimization behaviour, it is frequently hard to assess which emotions should be elicited by the agent at each point in the world. On the other hand, more realistic scenarios quickly become high-dimensional, and therefore the challenge changes to a representation learning problem. Potentially, the advances in solving more complex AI scenarios with (deep) RL \citep{silver2016mastering,mnih2015human} may provide more realistic test scenarios in the future as well. 

There are two other important observations regarding testing and evaluation. We have not encountered any (emotional) scenario being reproduced by other researchers. This appears to us as an important problem. To enhance the standard of the field, researchers should start reproducing scenarios from other's work to compare with, or borrow from different RL literature. The second topic we want to emphasize is the use of different evaluation criteria. Researchers should choose whether they target learning efficiency, emotion dynamics or HRI criteria. If learning performance is your criterion, then your implementation must include a baseline. When you focus on emotion dynamics, then you should try to validate by a psychological theory, or ideally compare to empirical (human) data. When you focus on human interaction criteria, then this should usually involve a questionnaire. Although questionnaires seems to be consistent practice already, we did observe authors reporting on a smaller subset of the questions (i.e. posing the risk to have a few results pop out by statistical chance).

This brings us to a final problem in the field, being the thoroughness of the papers. Frequently we were unable to fully deduce the details of each implementation. Indeed a full system description with all the details requires valuable space, but on the other hand, a well-informed colleague reading a conference paper should be able to reproduce your results. Only listing the homeostatic/appraisal variables and the emotions that were implemented does not provide deeper understanding about how the system works. This also makes it harder to compare between implementations. Differences in notational conventions and slight differences in definitions further complicate comparisons. Paying attention to these aspects of reproducibility, for example sticking to conventional RL notation \citep{sutton1998reinforcement}, will facilitate broader uptake of the work in this field. 

\paragraph{Future.}
A core challenge for the future will be to integrate all aspects into one larger system, potentially taking a fully learned approach. Along the same line, it is a remaining challenge of this field (and AI in general) to translate higher-level (psychological) concepts into implementable mathematical expressions. Examples of such translations can be found in Equations \ref{ab1}-\ref{ab5}, and we expect comparing different translations may help identify more consensus. At least the RL framework provides a common language to start comparing these translations.   

With social robots increasingly positioned at our research horizon, we expect interest in emotion in functional agents to increase in the forthcoming years. However, the current implementations seldomly investigate the full social interaction. Although this is a very high-level AI challenge, we believe research should focus in this direction to show empirical success. This involves all aspects of RL in a social context, i.e. robots learning from human demonstration (LfD) \citep{argall2009survey}, learning from human feedback (possibly emotional \citep{broekens2007emotion}), human emotions influencing agent emotions, and agent emotions communicating internal processes back to humans.

From an affective modelling perspective, it is promising to see how a cognitive theory like appraisal theory turns out to be well-applicable to MDP settings. Apart from integrating important lines of emotion and learning research, this also illustrates how cognitive and learning theories are not mutually exclusive. We hope the affective modelling community will start to benefit from the literature on intrinsic motivation in RL as well \citep{bratman2012strong}. A crucial requisite herein will be improving the types of problems that (model-based) RL can solve. Many scenarios that are interesting from an affective modelling viewpoint, for example high-dimensional social settings, are still challenging for RL. Advances in deep reinforcement learning \citep{mnih2015human} might make more complex scenarios available soon. However, for affective modelling we especially need the transition function and model-based RL \citep{deisenroth2011pilco}. Recent work has also shown the feasibility of high-dimensional transition function approximation \citep{oh2015action} in stochastic domains \citep{moerland2017learning} under uncertainty \citep{houthooft2016curiosity}. Further progress in this direction should make the ideas covered in this survey applicable to more complicated scenarios as well. 

\section{Conclusion}
This article surveyed emotion modelling in reinforcement learning (RL) agents. The literature has been structured according to the intrinsically motivated RL framework. We conclude by identifying the main benefits encountered in this work for the machine learning (ML), human-robot interaction (HRI), and affective modelling (AM) communities. For machine learning, emotion may benefit learning efficiency by providing inspiration for intrinsic motivation, exploration and for meta-parameter tuning. The current results should stimulate further cross-over between (intrinsic) motivation, model-based RL and emotion-RL research. For HRI research, emotions obviously are important for social interaction. More work should be done on implementing emotion models in interactive reinforcement learning algorithms, for which the survey presents a practical guideline on implementing emotions in RL agents. For affective modelling, we conclude that cognitive theories (like appraisal theory) can well be expressed in RL agents. The general benefits of RL agents (they require little assumptions, are easily applicable to all kinds of domains, and allow for learning) make them a promising test-bed for affective modelling research. This survey identifies opportunities for future work with respect to implementation and evaluation of emotion models in RL agents.  

%% Addendum: Large Overview Tables
\clearpage
{\scriptsize
\begin{longtable}{p{1.2cm} p{3.1cm} p{3cm} p{3.1cm}} 
\caption{Systematic overview of emotion elicitation, emotion type and emotion function in the reinforcement learning loop (see Figure \ref{schema}). Papers are ordered by their elicitation method (first column). Note that for homeostatic specification, we try to use the terms mentioned in the original paper, which may sometimes refer to the drive (i.e. the deficit in homeostatic variable) rather than the homeostatic dimension itself. Colour coding is based on the first term mentioned in each cell, grouping the categories as encountered in Sections \ref{elicitation}-\ref{function} and Table \ref{overviewtable}.}  \label{overview1} \\

\hline \noalign{\smallskip}
\bf Paper & {\bf Emotion Elicitation} & {\bf Emotion Type} & {\bf Emotion Function} \\
\noalign{\smallskip} \hline \noalign{\smallskip}

\citep{gadanho1998emotion,gadanho2001robot} &   Homeostasis: hunger, pain, restlessness, temperature, eating, smell, warmth, proximity \cellcolor{blue!15} & Categorical: happiness, sadness, fear, anger & Reward modification: positive emotion is reward \cellcolor{orange!15} \\

\hline
\citep{gadanho2003learning} &   Homeostasis: energy, welfare, activity \cellcolor{blue!15} & Dimensional: well-being \cellcolor{black!10} & Reward modification: delta well-being is reward \cellcolor{orange!15} \\

\hline
\citep{cos2013hedonic} & Homeostasis: hunger, tiredness, restlessness \cellcolor{blue!15}  & Categorical: hedonic value & Reward modification: delta well-being is reward \cellcolor{orange!15} \\

\hline
\citep{coutinho2005towards} & Homeostasis: blood sugar, energy, pain, vascular volume, endorphine \cellcolor{blue!15}  & Dimensional: wellness, relaxation, fatigue \cellcolor{black!10} & Epiphenomenon \cellcolor{cyan!15}  \\

\hline
\citep{von2012modulating} &   Homeostasis: hunger, fatigue, interest {\cellcolor{blue!15}} & Categorical: happiness, sadness, anger, surprise, fear, disgust. & Metalearning: reward = delta happiness, learning rate = (1-surprise), discount factor = (1-fear), Boltzmann temperature = anger \cellcolor{brown!15}  \\

\hline
\citep{tanaka2004associated} &   Homeostasis: hunger, fullness, pain, comfort, fatigue, sleepiness {\cellcolor{blue!15}} & Categorical: happiness, sadness, anger, surprise, disgust, fear, neutral & Epiphenomenon: gesture, voice, facial expression \cellcolor{cyan!15} \\

\hline
\citep{goerke2006emobot} &   Homeostasis: fatigue, hunger, homesickness, curiosity \cellcolor{blue!15} & Categorical: happiness, fear, anger, boredom & Reward modification: positive emotion is reward \cellcolor{orange!15} \\

%%%
\hline
\citep{sequeira2011emotion,sequeira2014learning} & Appraisal: valency, control, novelty, motivation \cellcolor{green!15}  & None   & Reward modification: summed appraisals added to reward function \cellcolor{orange!15} \\

\hline
\citep{marinier2008emotion} & Appraisal: suddenness, intrinsic pleasantness, relevance, conduciveness, discrepancy from expectation, control, power. \cellcolor{green!15}&  None  & Reward modification: summed appraisals is reward \cellcolor{orange!15} \\

\hline
\citep{yu2015emotional,yu2013emotional} & Appraisal: social fairness \newline Value: average reward \cellcolor{green!15} & Categorical: happiness, sadness, fear, anger & Reward modification: positive/negative emotion is positive/negative reward \cellcolor{orange!15} \\

\hline
\citep{lee2010mobile,lee2007emotion} & Appraisal: model mismatch \cellcolor{green!15} \newline Value: average achieved reward, global planned reward \newline Homeostatic: collision & Categorical: Happiness, sadness, fear, anger, surprise & Reward modification: change local reward (happy \& suprise higher, fear \& anger lower)  \cellcolor{orange!15} \\

\hline
\citep{williams2015emotion} & Appraisal: novelty, progress, control, uncertainty \cellcolor{green!15} \newline Homeostatic: pain. & Categorical: happiness, fear (post-characterized) & Metalearning: happy gives positive reward bias and higher travel speed, fear giver negative reward bias and lower travel speed \cellcolor{brown!15}  \\

\hline
\citep{si2010modeling} & Appraisal: motivational relevance \& congruence, accountability, control, novelty. \cellcolor{green!15} & None  & Epiphenomenon \cellcolor{cyan!15} \\

\hline
\citep{kim2010computational} & Appraisal: unexpectedness, motive consistency, control, uncertainty, agency/accountability.. \cellcolor{green!15} & Dimensional: valence, arousal (not fully explicit)  & Epiphenomenon: facial avatar, voice, movement of ears, music \cellcolor{cyan!15} \\

\hline
\citep{hasson2011emotions} & Appraisal: non-progress \newline Human affective state \cellcolor{green!15} & Categorical: frustration, anger, happiness & Action selection: switch between targets \cellcolor{yellow!20}  \\

\hline
\citep{moussa2013toward} & Appraisal: desirability, attachment (OCC model) \cellcolor{green!15} & Categorical: joy, distress, happy for, resentment, sorry for, gloating, gratitude, admiration, anger, reproach. & Reward modification: reward is difference of largest positive and negative current emotion \cellcolor{orange!15} \\

\hline
\citep{huang2012goal} & Appraisal: motivational relevance + goal reachable \cellcolor{green!15} & Categorical: Happy, sad, anger, surprise, fear, frustration & Epiphenomenon \cellcolor{cyan!15} \\

\hline
\citep{kuremoto2013improved} & Appraisal: distance to goal, continuation of eliciting event \cellcolor{green!15} & Dimensional: valence, arousal \cellcolor{black!10}  & Action selection: separate emotional Q-value as part of total summed Q-value \cellcolor{yellow!20} \\

%%%%%%%%%%
\hline
\citep{castro2013autonomous,salichs2012new} & Value: worst historical Q-value + \newline Homeostasis: energy,  boredom, calm, loneliness \cellcolor{red!15} & Categorical: happiness, sadness, fear & Reward modification: delta well-being \newline State modification: fear replaces dominant motivation (when threshold is exceeded)  \cellcolor{orange!15} \\

\hline
\citep{ahn2006affective} & Reward: difference between experienced reward and expected immediate reward of best two available actions \cellcolor{red!15} & Dimensional: feeling good, bad \cellcolor{black!10} &  Action selection: emotional Q-value is part of total Q-value  \cellcolor{yellow!20}  \\

\hline
\citep{zhang2009design}  & Reward: difference between experienced reward and expected immediate reward of best action \cellcolor{red!20}& Dimensional: feeling good/bad \cellcolor{black!10}& Action selection: emotional Q-value is part of total Q-value \cellcolor{yellow!20}   \\

\hline
\citep{broekens2007affect,broekens2007exploration} & Reward: short versus long term average reward \cellcolor{red!15} & Dimensional: valence \cellcolor{black!10}  & Action selection: emotion tunes exploration parameter and simulation depth \cellcolor{yellow!20} \\

\hline
\citep{moerland2016hope} & Value: Anticipated temporal difference \cellcolor{red!15} & Categorical: hope, fear  & Epiphenomenon \cellcolor{cyan!15} \\

\hline
\citep{jacobs2014emergent} & Value: temporal difference and positive/negative part of value \cellcolor{red!15} & Categorical: joy, distress, hope, fear & Epiphenomenon \cellcolor{cyan!15} \\

\hline
\citep{bozinovski1982self} & Value \cellcolor{red!15} & None & Epiphenomenon \cellcolor{cyan!15} \\

\hline
\citep{moren2000computational} & Value \cellcolor{red!15} & None & Epiphenomenon \cellcolor{cyan!15} \\

\hline
\citep{lahnstein2005emotive} & Value: temporal difference \cellcolor{red!15} & Categorical: happiness, sadness, hope. & Epiphenomenon \cellcolor{cyan!15} \\

\hline
\citep{obayashi2012emotional}  & Reward: not explicit \newline Hard-wired: not-explicit \cellcolor{red!15} & Dimensional: valence, arousal (with unlabelled categories) \cellcolor{black!10} & State modification: emotion specific Q-value \cellcolor{purple!15}  \\

\hline
\citep{matsuda2011decision} & Reward: only negative reward \cellcolor{red!15} & Categorical: fear & Action selection: separate emotional value function is part of action selection \cellcolor{yellow!20} \\

\hline
\citep{schweighofer2003meta,doya2002metalearning} & Reward: mid versus long-term average reward  \cellcolor{red!15} & None & Metalearning: perturbation of discount, learning and temperature parameter based on emotion. \cellcolor{brown!15}  \\

\hline
\citep{hogewoning2007strategies} & Reward: short/mid versus long-term average reward   \cellcolor{red!15} & Dimensional: valence \cellcolor{black!10} & Action selection: emotion tunes exploration (combines \citep{broekens2007exploration} and \citep{schweighofer2003meta} with chi-square test) \cellcolor{yellow!20} \\

\hline
\citep{shi2012artificial} & Reward: change in reward signal \cellcolor{red!15} & Categorical: joy, fear, anger, relief & Metalearning: joy = TD, anger = temperature, fear = learning rate, relief = discount parameter (connection not explicit) \cellcolor{brown!15}  \\

\hline
\citep{blanchard2005imprinting} & Reward: average \cellcolor{red!15} & Categorical: comfort & Metalearning: emotion modulates the learning rate \cellcolor{brown!15}  \\

\hline
\citep{el2000flame} & Value: combined with fuzzy logic \cellcolor{red!15} & Categorical: joy, sadness, disappointment, relief, hope, fear, pride, shame, reproach, anger, gratitude, gratification, remorse & Action selection: emotions are input to a fuzzy logic action selection system \cellcolor{yellow!20} \\

%%%
\hline
\citep{kubota2010emotional}& Hard-wired: from objects (users, balls, chargers, obstacles), speech and previous emotional state \cellcolor{yellow!20} & Categorical: happiness, sadness, fear, anger & Action selection: switch between value functions \cellcolor{yellow!20} \\

\hline
\citep{ayesh2004emotionally} & Hard-wired: from state through fuzzy cognitive maps \cellcolor{yellow!20}& Dimensional: restless, neutral, stable \cellcolor{black!10} & State modification: emotion specific Q-values \cellcolor{purple!15}  \\

\hline
\citep{ficocelli2015promoting} &   Human affective state \newline Hard-wired  \cellcolor{yellow!20} & Categorical: happiness, neutral, sadness, angry. & State modification \cellcolor{purple!15} \newline Action selection: modify intonation of speech \\

\hline
\citep{hoey2015bayesian} & Hard-wired from object observations (social interaction) \cellcolor{yellow!20}& Dimensional: valence, control, arousal \cellcolor{black!10} & State modification: extended POMDP derivation with 3D emotional state \cellcolor{purple!15}   \\

\hline
\citep{tsankova2002emotionally} & Hardwired: from obstacle detectors \cellcolor{yellow!20}& Categorical: frustration & Action selection: emotion controls the balancing between value functions \cellcolor{yellow!20} \\

\hline
\citep{zhou2002computational} & Hardwired: from sight of resources \cellcolor{yellow!20} \newline Homeostasis: hunger, thirst (not connected to emotion but to reward) & None & Reward modification: reward calculated from maximum emotion or motivation. \cellcolor{orange!15} \\

\hline
\citep{doshi2004towards} & Hard-wired: from sight of enemy or resource. \cellcolor{yellow!20} & Categorical: Contented, elation, fear, panic. & Action selection: emotion adjust planning depth and biases considered actions \cellcolor{yellow!20} \\

\hline
\citep{gmytrasiewicz2002emotions} & Hard-wired: Markovian transition from previous emotions and state \cellcolor{yellow!20} & Categorical: Cooperative, slightly annoyed, angry & Meta-learning: emotion biases transition function \newline Action selection: emotion biases available action subset, biases value function \cellcolor{brown!15} \\

\hline
\citep{guojiang2010behavior} & Hard-wired: from exterior incentive like safety, threat, fancy, surprise (assumed pre-given) \cellcolor{yellow!20} & Dimensional: valence, arousal \cellcolor{black!10} & State modification: 2D emotional state space \newline Reward modification: agent should move to desirable area in emotional space (implementation not specified) \cellcolor{purple!15} \\

\hline
\citep{shibata1997artificial} & Not explicit & Not explicit & Not explicit \\

\noalign{\smallskip} \hline
%\end{tabular}
\end{longtable}
}

%%%%%%%%%%%%%%%%%%%% Table 2 %%%%%%%%%%%%%%%%%%%%%%
%%%%%%%%%%%%%%%%%%%%%%%%%%%%%%%%%%%%%%%%%%%%
%%%%%%%%%%%%%%%%%%%%%%%%%%%%%%%%%%%%%%%%%%%%

{\scriptsize
\begin{longtable}{p{1.3cm} p{1.2cm} p{1.4cm} p{1.1cm} p{5cm} } 
\caption{Systematic overview of test embodiment, scenario, evaluation criterion and main results. Papers are ordered according to Table \ref{overview1}. Colour coding presented for the evalution criterion column.} \label{overview2} \\
\hline \noalign{\smallskip}
\bf Paper & {\bf Embodi-ment} &{\bf Scenario} & {\bf Criterion} & {\bf Main result} \\
\noalign{\smallskip} \hline \noalign{\smallskip}

\citep{gadanho1998emotion,gadanho2001robot,gadanho2003learning} & Simulated robot & Multiple resource task & Learning \cellcolor{yellow!20} & Less collisions and higher average reward with emotional agent.  \\

\hline
\citep{cos2013hedonic} & Grid-world agent & Multiple resource task & Learning \cellcolor{yellow!20} & Emergent behavioural cycles fulfilling different drives   \\

\hline
\citep{coutinho2005towards} & Grid-world agent & Multiple resource task  & -  & No emotion results \\

\hline
\citep{von2012modulating} & Multiple agents & Game / competition & Learning \cellcolor{yellow!20} & Increased average reward compared to non-emotional agents \\

\hline
\citep{tanaka2004associated} & Real robot & Human interacting ( hitting/ padding robot) & \cellcolor{blue!20}Dynamics& Appropriate emotion response (fear and joy) to bad and good acting person.   \\

\hline
\citep{goerke2006emobot} & Simulated robot + real robot & Multiple resource task & Learning \cellcolor{yellow!20} & Different behaviour types with emotion functionality \\

%%%
\hline
\citep{sequeira2011emotion} & Grid-world agent & Resource-predator task  & Learning \cellcolor{yellow!20} & Improved average fitness compared to non-appraisal agent \\

\hline
\citep{marinier2008emotion} & Grid-world agent & Maze & Learning \cellcolor{yellow!20} & Emotional agent needs less learning episodes \\

\hline
\citep{yu2015emotional,yu2013emotional} & Multiple agents & Game / negotiation  & Learning \cellcolor{yellow!20} & Emotional/social agents have higher average reward and show co-operation  \\

\hline
\citep{lee2010mobile,lee2007emotion} & Simulated robot &  Navigation task  & Learning \cellcolor{yellow!20} & Emotional agent has less collisions and more exploration, against a higher average travel time \\

\hline
\citep{williams2015emotion} & Real robot & Navigation task  & Learning \cellcolor{yellow!20} & Less collisions with fear enabled, more exploration with surprise, quicker routes with happiness enabled.\\

\hline
\citep{si2010modeling} & Multiple agents & Social interaction &  Dynamics \cellcolor{blue!20} & Different appraisal with deeper planning + Social accountability realistically derived (compared to other computational model). \\

\hline
\citep{kim2010computational} & Real robot & Social interaction (question game) with human & HRI \cellcolor{green!20} & Users report higher subjective feeling of interaction and higher pleasantness for emotional robot + humans correctly identify part of the underlying robot appraisals based on a questionnaire \\

\hline
\citep{hasson2011emotions} & Real robot & Multiple resource navigation task & Learning \cellcolor{yellow!20} & Robot with emotion can switch between drives (in case of obstacles) and escape deadlocks.  \\

\hline
\citep{moussa2013toward} & Real robot & Human dialogue task (while playing game) & Dynamics + Learning \cellcolor{blue!20} & Appropriate emotion responses to friendly and unfriendly users + learn different attitudes towards them.   \\

\hline
\citep{huang2012goal} & Grid-world agent & Navigation task & Dynamics \cellcolor{blue!20} & Dynamics show how emotion elicitation varies with planning depth and goal achievement probability.  \\

\hline
\citep{kuremoto2013improved} & Grid-world agent & Predator task & Learning \cellcolor{yellow!20} & Quicker goal achievement for emotional agent compared to non-emotional agent. \\

%%%%%%%%%%
\hline
\citep{castro2013autonomous,salichs2012new} & Real robot & Multiple resources task including human objects & Learning + Dynamics \cellcolor{yellow!20} & Less harmful interactions compared to non-fear robot + realistic fear dynamics (compared to animal) \\

\hline
\citep{ahn2006affective} & Agent & Conditioning experiment & Learning \cellcolor{yellow!20} & Affective agent learns optimal policy faster. \\

\hline
\citep{zhang2009design} & Simulated robot & Navigation task & Learning \cellcolor{yellow!20}  & Emotional robot needs less trials to learn the task.  \\

\hline
\citep{broekens2007affect,broekens2007exploration} & Grid-world agent & Maze & Learning \cellcolor{yellow!20} & Emotional control of simulation depth improves average return. Emotional control of exploration improves time to goal and time to find the global optimum.  \\

\hline
\citep{moerland2016hope} & Grid-world agent + Pacman & Resource-predator task & Dynamics \cellcolor{blue!20}  & Appropriate hope and fear anticipation in specific Pacman scenarios. \\

\hline
\citep{jacobs2014emergent} & Grid-world agent & Maze & Dynamics \cellcolor{blue!20} & Emotion dynamics (habituation, extinction) simulated realistically compared to psychological theory. \\

\hline
\citep{bozinovski1982self,bozinovski1996emotion} & Grid-world agent & Maze & Learning \cellcolor{yellow!20} & First investigation of emotion as primary reward, shows agent is able to solve maze task.  \\

\hline
\citep{moren2000computational,balkenius1998computational} & Agent & Conditioning experiment  & Dynamics \cellcolor{blue!20} & Agent shows habituation, extinction, blocking (i.e. of learning signal, not emotion) \\

\hline
\citep{lahnstein2005emotive} & Real-robot & Multiple objects grasping task & Dynamics \cellcolor{blue!20} & Models dynamics within single decision cycle, shows plausible anticipation, hedonic experience and subsequent decay. \\

\hline
\citep{obayashi2012emotional} & Grid-world agent & Maze & Learning \cellcolor{yellow!20} & Emotional agent needs less steps to goal (ordinary agent does not converge). \\

\hline
\citep{matsuda2011decision} & Multiple agent grid-world & Co-operation task & Learning \cellcolor{yellow!20} & Emotional agents show more co-operation and adapt better to environmental change compared to non-emotional agents.  \\

\hline
\citep{schweighofer2003meta,doya2002metalearning} & Agent & Conditioning experiment + Simulated pendulum & Learning \cellcolor{yellow!20} & Dynamic adaptation of meta-parameters in both static and dynamic environment. Task not achieved for fixed meta-parameters.   \\

\hline
\citep{hogewoning2007strategies} & Grid-world agent & Maze & Learning \cellcolor{yellow!20}  & Emotional agent cannot improve results of \citep{broekens2007exploration,schweighofer2003meta} \\

\hline
\citep{shi2012artificial} & Grid-world agent & Obstacle and resource task & Learning + Dynamics \cellcolor{yellow!20} & Emotional agent avoids obstacle better. Different emotion lead to different paths. \\

\hline
\citep{blanchard2005imprinting} & Real robot & Conditioning task &  Dynamics \cellcolor{blue!20} & Robot can imprint desirable stimuli based on comfort (reward) signal, and subsequently show approach or avoidance behaviour.  \\

\hline
\citep{el2000flame} & Screen agent & Human interaction task & HRI \cellcolor{green!20} & Users perceive agent with emotional action selection as more convincing.  \\

%%%
\hline
\citep{kubota2010emotional} & Simulated robot & Multiple objects and human & Dynamics \cellcolor{blue!20} & Emotional robot avoids dangerous areas due to fear, and starts exploring when happy. \\

\hline
\citep{ayesh2004emotionally} & Real robot & None & None &  None \\

\hline
\citep{ficocelli2015promoting} & Real robot & Human dialogue task & HRI + Dynamics + Learning \cellcolor{green!20}  & Effective emotion expression (user questionnaire) + Robot changing emotions to satisfy different drives \\

\hline
\citep{hoey2015bayesian} & Agent & Social agent interaction & Dynamics \cellcolor{blue!20} & Model can accurately modify own dimensional emotion with respect to the client it is interacting with.  \\

\hline
\citep{tsankova2002emotionally} & Simulated robot & Navigation task.  & Learning \cellcolor{yellow!20} & Emotional robot reaches goal more often, but need more timesteps. \\

\hline
\citep{zhou2002computational} & Real robot & Multiple resources & Learning \cellcolor{yellow!20}  & Emotional robot has higher average reward and less intermediate behaviour switching compared to non-emotional robot. \\

\hline
\citep{doshi2004towards} & Grid-world & Multiple resource, predator task & Learning \cellcolor{yellow!20} & Emotional agent (with meta-learning) has higher average return compared to non-emotional agent. \\

\hline
\citep{gmytrasiewicz2002emotions} & None & None (theoretical model) & None & None \\

\hline
\citep{guojiang2010behavior} & Agent & Conditioning task & Dynamics \cellcolor{blue!20} & Agent moves towards beneficial emotional state-space and stays there. \\

\hline
\citep{shibata1997artificial} & Real robot & Human stroke/pad robot & HRI \cellcolor{green!20} & Humans reported a coupling with the robot, some reported it as intelligent. Subjects report positive emotions themselves. \\

\noalign{\smallskip} \hline
\end{longtable}

%\begin{acknowledgements}
%\end{acknowledgements}

%\bibliographystyle{apalike}
\bibliographystyle{spbasic}      % basic style, author-year citations
\bibliography{extracted.bib}  % name your BibTeX data base

\end{document}